\pdfoutput=1

\documentclass[11pt]{article}

\usepackage{acl}
\usepackage{graphicx}

\usepackage{times}
\usepackage{latexsym} 
\usepackage{subcaption}
\usepackage[font=Large]{subcaption}
\usepackage{amsmath}
\usepackage{amssymb}
\usepackage{amsthm}
\usepackage{comment}

\newtheorem{lemma}{Lemma}
\usepackage{multicol}
\usepackage{enumitem}

\usepackage[T1]{fontenc}

\usepackage[utf8]{inputenc}

\usepackage{microtype}

\usepackage{inconsolata}
\usepackage{mathtools}

\usepackage{graphicx} 
\usepackage{algorithm}
\usepackage{algorithmic}

 \usepackage{hyperref} 
\usepackage[arabic,english]{babel}

\usepackage[most]{tcolorbox}
\usepackage{placeins} 

\usepackage{booktabs}
\usepackage{multirow}
\usepackage{tabularx}
\usepackage{array}
\usepackage{bm}
\usepackage{siunitx}
\newcommand{\ourmethod}{{\textsc{Voyager}}} 

\usepackage{eqparbox}
\renewcommand{\algorithmiccomment}[1]{\hfill\eqparbox{COMMENT}{\textcolor{gray}{\# #1}}}

\title{\ourmethod: A Training Free Approach for Generating Diverse Datasets using LLMs}

\author{
  \textbf{Avinash Amballa} \hspace{5mm} \textbf{Yashas Malur Saidutta} \hspace{5mm}  \textbf{Chi-Heng Lin} \\ \\
  \textbf{Vivek Kulkarni} \hspace{5mm}  \textbf{Srinivas Chappidi} \\ \\
  \textbf{Samsung Research America} \\
  \texttt{\{a.amballa, ym.saidutta, chiheng.lin, v.kulkarni1, vasu.c\}@samsung.com} \\
}

\begin{document}
\maketitle
   
\begin{abstract}
Large language models (LLMs) are increasingly being used to generate synthetic datasets for the evaluation and training of downstream models. However, prior work has noted that such generated data lacks diversity. In this paper, we propose \ourmethod\ , a novel principled approach to generate diverse datasets. Our approach is iterative and directly optimizes a mathematical quantity that optimizes the diversity of the dataset using the machinery of determinantal point processes. Furthermore, our approach is training-free, applicable to closed-source models, and scalable. In addition to providing theoretical justification for the working of our method, we also demonstrate through comprehensive experiments that \ourmethod\ improves diversity by $\mathbf{1.5}$-$\mathbf{3}$ times compared to popular baseline approaches.
\footnote{\url{https://sites.google.com/view/avinashamballa/projects/voyager}}

\end{abstract}

\section{Introduction}

Large language models (LLMs) are widely used to generate synthetic data for scenarios where real world data is scarce. Although such data is valuable for training downstream models, post-trained LLMs used to generate such data often exhibit poor diversity in generation~\cite{wright2025epistemic, padmakumar2023does}. Common strategies to address this include sampling methods, such as temperature, top-$p$ sampling and min-$p$ sampling \cite{pmlr-v97-kool19a, holtzman2020curiouscaseneuraltext}. However, these operate on the next token distribution and often fail to prevent mode collapse or semantic clustering, even at high temperatures \cite{jiang2025artificialhivemindopenendedhomogeneity}; as they lack global view of diversity between generations. Alternatively, ``prompt based diversity control'' instructs LLMs to cover specific topics. Although applicable to closed-weight models, this approach is not scalable and requires significant domain expertise. Finally, while diversity can be encoded into post-training rewards~\cite{li2025jointly, chen2025post}, such methods are computationally expensive and restricted to open-weight models.

We introduce \ourmethod, a novel training-free framework for diverse synthetic data generation. Grounded in mathematical theory, our approach optimizes a geometric property intrinsic to dataset diversity. Crucially, \ourmethod\ operates without accessing model parameters, ensuring computational efficiency and compatibility with closed-weight (black-box) models. \ourmethod\ is inspired by cartography, where explorers guided by a central command are encouraged  to explore uncharted regions to maximize coverage. Analogously, our framework explores the data manifold, accepting generations that differ significantly from prior samples. To ensure computational efficiency, we maintain a fixed-size ``anchor set'' of representative regions rather than comparing against the full history. We dynamically prune this set to maximize its geometric volume, instructing the model to avoid clustered areas and expand diversity. 

\ourmethod\ formalizes this intuition using Determinantal Point Processes. Empirical evaluations show that \ourmethod\ consistently outperforms strong baselines across a suite of tasks.To summarize, our main contributions are:
\begin{itemize}[noitemsep]
\item \textbf{Principled approach to maximizing diversity:} Our approach leverages the machinery of determinantal point processes to maximize diversity by formulating it in terms of a geometric metric (volume). This provides theoretical justification to our approach.
\item \textbf{Applicable to both open and closed weight models:} Our approach is training-free and requires no access to model weights.
\item \textbf{Strong empirical performance:} We improve diversity by $1.5-3$ times over baselines.


\end{itemize}

\section{Related Work}

Existing diversity promotion methods fall into two categories: training-free or training-based.

\subsection{Training-free Methods}
These methods do not update model weights and use  decoding strategies or prompt engineering.

\paragraph{Sampling and Decoding Strategies} 
Some approaches manipulate output probabilities and decoding strategies via techniques like temperature, nucleus, min-$p$, or top-$p$ \cite{pmlr-v97-kool19a, holtzman2020curiouscaseneuraltext, amballa2025quasirandom, hewitt2022truncation, minhturning, chang2024real}, or utilize diversity-rewarding beam search~\cite{vijayakumar2016diverse}. However, these methods require access to raw logits, making them unsuitable for closed-source models and are prone to incoherent or grammatically flawed outputs. In contrast, our method optimizes a global diversity proxy without requiring access to model weights or logits.

\paragraph{Prompt Diversity Control} 
Prompt-based methods guide generation by explicitly instructing the model to vary dimensions such as topic or style~\cite{wong2024simplestrat, samvelyan2024rainbow}; or refine outputs across multiple turns \cite{mehrotra2024enhancing, tian2024macgyver, lu2025benchmarking, wong2024simplestrat}. While effective for open and closed models, these approaches rely on pre-defined and granular topics. Consequently, they optimize for topical diversity rather than semantic diversity; generations from distinct topics may still share significant vocabulary. Unlike these heuristics, our method directly maximizes the semantic volume of the data, ensuring diversity beyond topical differences.

\subsection{Training-based Methods}
Training-based approaches fine-tune model weights to maximize diversity-centric reward functions~\cite{li2025jointly, yao2025diversity, chen2025post, ismayilzada2025creative, lanchantin2025diverse}. Although effective, these methods incur significant computational overhead and are inapplicable to closed-source models. 

Closely related to our work, \citet{wang2024diversity} propose using volume to measure dataset diversity for selection purposes. Similarly, \citet{chen2025post} integrate the determinant of the similarity matrix into post-training.\footnote{While \citet{chen2025post} also utilize a determinant-based metric, their approach is a weight-tuning method restricted to open models and focuses on local response diversity. We focus on generating a globally diverse dataset via a training-free iterative algorithm.} However, whereas these works focus on \textit{measuring} diversity or \textit{aligning} models, we introduce a novel, iterative algorithm that leverages ``text-based gradients'' to actively \textit{generate} a globally diverse dataset.

\begin{figure*}[t] 
    \centering 
    \includegraphics[trim={1cm 5cm 4cm 0cm },clip, width=\textwidth]{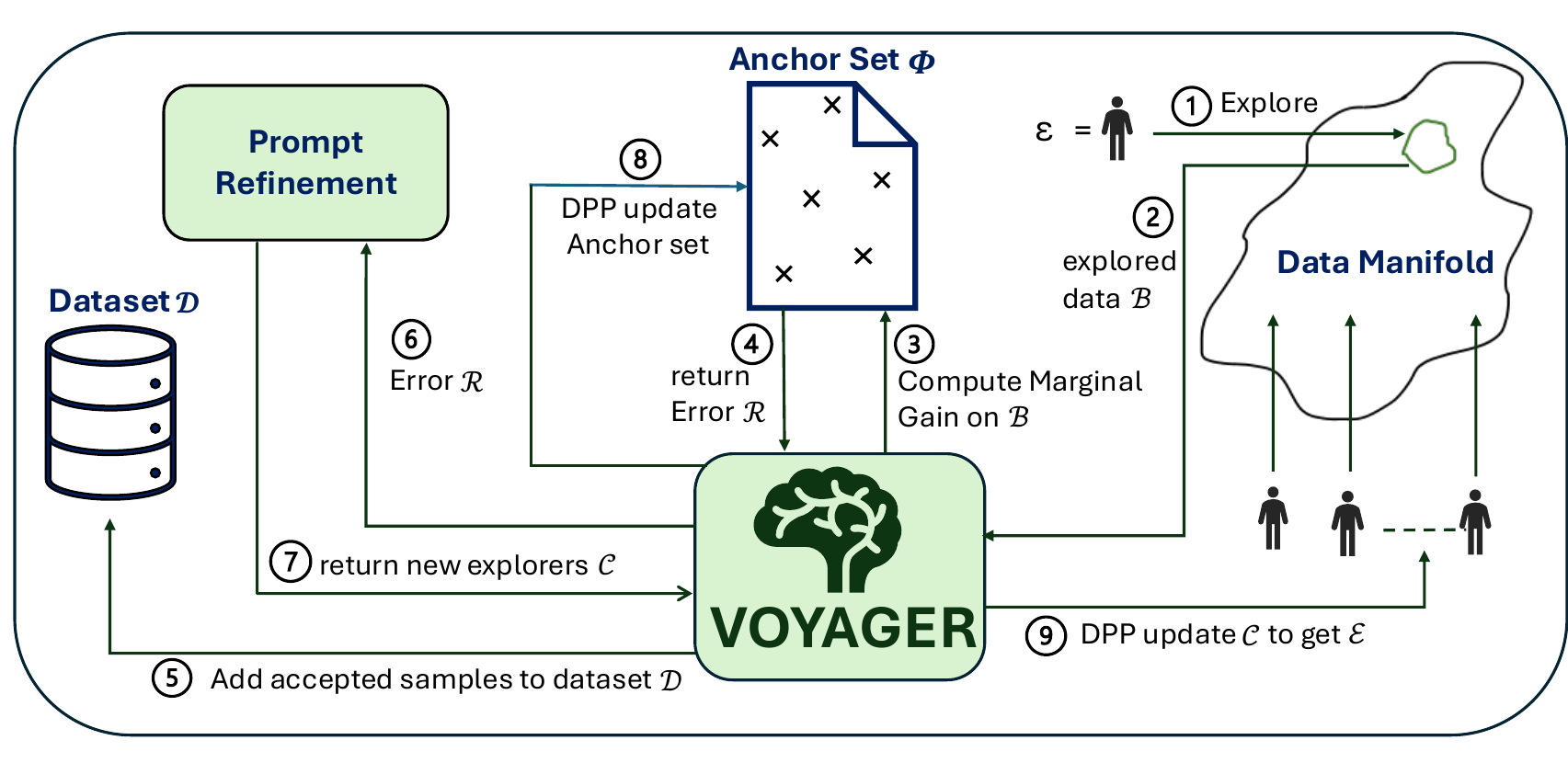}
    \caption{\textbf{Overview of \ourmethod.} We iteratively seek to explore new diverse regions of the data manifold via set of successive voyages carried out by explorers.  Each explorer explores a certain region of the manifold (steps 1-2). Regions that are very similar to prior explorations are rejected by the central command which keeps track of a key set of salient regions explored, i.e., the anchor set (steps 3-5). New explorers are encouraged to explore areas different from prior explorations via prompt refinement (steps 6-9).}
    \label{fig:overview} 
\end{figure*}

\section{Background}
\ourmethod\ relies on two key pieces of machinery: (a) Determinantal Point Processes (b) Prompt refinement using ``textual'' gradients. 

\noindent \textbf{Determinantal Point Processes:} A Determinantal Point Process (DPP) \cite{kulesza2012determinantal} is a probabilistic model for subset selection where the probability of selecting a subset $S$ is proportional to the determinant of the corresponding kernel matrix. For a ground set $\mathcal{Y}$ and a positive semi-definite kernel matrix $K \in \mathbb{R}^{|\mathcal{Y}| \times |\mathcal{Y}|}$, the probability of selecting subset $S \subseteq \mathcal{Y}$ is $\mathbb{P}(S \sim \text{DPP}(K)) \propto \det(K_S)$, 
where $K_S$ is the submatrix of $K$ indexed by elements in $S$. The determinant $\det(K_S)$ measures the \emph{volume} spanned by the vectors corresponding to items in $S$, directly encoding diversity: higher determinants (volumes) correspond to more diverse subsets with less redundancy. 

\paragraph{Prompt Refinement using Textual Gradients} This is a prompt refinement procedure (analogous to gradient descent, but operates purely in the text space) introduced by ~\citet{pryzant2023automaticpromptoptimizationgradient}. The objective here is to improve the prompt with respect to some reward metric. In the first of the two steps, an LLM-judge is asked to judge the prompt and its corresponding generation and suggest changes to the prompt so that the reward metric might be increased. In the second step, these suggestions are provided to another LLM is tasked to incorporate them into the prompt. 

\section{\ourmethod\ -- Algorithm}


\ourmethod\ mathematically operationalizes the notion of diversity of a set by the determinant of the corresponding similarity matrix of the set (encoded using a suitable kernel) and seeks to approximately maximize this measure iteratively.  

\begin{algorithm}[t]
\caption{\ourmethod}
\label{alg:voyager}
\begin{algorithmic}[1]
\REQUIRE $p$:task prompt \newline
$l$:desired dataset size \newline
$\tau$: Marginal gain threshold \newline
$b$:max number of explorers in each step \newline
$k$:maximum size of anchor point set \newline
$T$: Maximum number of iterations \\
$\mathbf{K}_\text{Sim}$: Similarity Kernel 
\ENSURE $\mathcal{D}$: Constructed dataset
\STATE $\mathcal{D} = \{\}$ \algorithmiccomment{Initialize dataset} 
\STATE $\Phi = \{\}$ \algorithmiccomment{Initialize anchor set} 
\STATE $\mathcal{E} = \{p\}$  \algorithmiccomment{Initialize set of explorers} 
\FOR{$i \leftarrow 0$ to $T$}
    \STATE $\mathcal{C} \leftarrow \{\}$ \algorithmiccomment{Successor set}
    \FORALL{$e \in \mathcal{E}$}
        \STATE $\mathbf{d}, \mathbf{A}, \mathbf{s} = \textsc{Explore}(e, \Phi, \tau, \mathbf{K}_\text{Sim})$ 
        \STATE $\mathcal{D} \leftarrow \mathcal{D}\cup\ \mathbf{d}$
        \STATE $\mathcal{C} \leftarrow \mathcal{C}\cup\ \mathbf{s}$
        \STATE $\Phi \leftarrow \textsc{SampleDPP}(\mathbf{A}, k, \mathbf{K}_\text{Sim})$
        \IF{$|\mathcal{D}| \geq l$}
         \STATE return $\mathcal{D}$
         \ENDIF
    \ENDFOR
    \STATE $\mathcal{E} \leftarrow \textsc{SampleDPP}(\mathcal{C}, b, \mathbf{K}_\text{Sim})$
\ENDFOR
\RETURN $\mathcal{D}$
\end{algorithmic}
\end{algorithm}

\begin{algorithm}[htb]
\caption{$\textsc{Explore}(\cdot)$}
\label{alg:explore}
\begin{algorithmic}[1]
\REQUIRE  $e$: explorer\newline
$\Phi$: Anchor Set\newline
$\tau$: Marginal Gain Threshold \newline
$\mathbf{K}_\text{Sim}$: Similarity Kernel 
\ENSURE $\mathcal{S}$: Data instances\newline
        $\Phi$: Augmented set of potential anchors\newline
        $\mathcal{C}$: Successor explorers
\STATE Prompt an LLM using $e$ to generate $\mathcal{B}$, a batch of instances (1 LLM call). 
\STATE $\mathcal{C} \leftarrow \{\}$ \algorithmiccomment{Successor set}
\STATE $\mathcal{S} \leftarrow \{\}$ \algorithmiccomment{Selected instances}
\STATE $\mathcal{R} \leftarrow \{\}$ \algorithmiccomment{Hold rejected instances}
\FORALL{$w \in \mathcal{B}$}
    \IF {\textsc{MarginalGain}(w, $\mathbf{K}_\text{Sim}$, $\Phi$) $\ge \tau$}
        \STATE Add $w$ to $\mathbf{S}$ and to $\Phi$
    \ELSE
        \STATE Add $w$ to rejected set $\mathcal{R}$ 
    \ENDIF
\ENDFOR

\IF{$|\mathcal{R}| > 0$}
\STATE Prompt an LLM to Get ``gradients'': $g = \{g_1, ..., g_m\} = \textsc{LLM}_{\nabla}(e,\mathcal{R}, \Phi)$ (1 LLM call)
\STATE Prompt an LLM to Apply gradients on $e$ to create a new set of explorers $\mathcal{C}$ (1 LLM call)
\ENDIF
\STATE return $\mathcal{S}$, $\Phi$, $\mathcal{C}$
\end{algorithmic}
\end{algorithm}




\subsection{Overview}
At a high level, our method (see Algorithm \ref{alg:voyager}, Figure \ref{fig:overview}) requires a task prompt $p$ that describes the specific data generation task (eg, Generate a poem). The algorithm maintains a fixed-size set of explorers ($\mathcal{E}$) and a fixed-size set of anchor data-points denoted by $\Phi$ that capture representative/diverse regions of the full underlying dataset. In each iteration of the outer loop, we pick the next explorer in the list and let the explorer perform a ``Explore'' (Line 7). This returns a new batch of data $\mathbf{d}$, an augmented set of potential anchors $\mathbf{A}$, and a successor set of candidate explorers $\mathbf{s}$. We add this new batch of data to the partially constructed dataset $\mathcal{D}$ and also add the set of newly generated candidate explorers to $\mathcal{C}$ (Lines 8-9). Because \ourmethod\ maintains a fixed-size set of anchor data-points to capture representative and diverse regions, and the newly added data-points could be potential anchor points, we update the anchor set $\Phi$ by sampling $k$ diverse points from the augmented set $\mathbf{A}$ (Line 10). Finally, after the current ``beam of explorers'' is done, we have a new set of candidate explorers $\mathcal{C}$. Once again, for computational efficiency, we select at most $b$ explorers for the next iteration. Because we favor diverse explorers, we once again sample $b$ explorers from $\mathcal{C}$ from the underlying DPP (see Algorithm \ref{alg:SampleDPP}) and use that sampled set in the next iteration (Line 15).
\subsubsection{Exploration Step}
Our \textsc{Explore} (see Algorithm \ref{alg:explore}) procedure consists of three main steps:
First, given an explorer, we generate a batch of data by calling an LLM with the prompt characterizing that explorer. 

We then score each data instance on its marginal gain in volume if it would be added to the anchor set. Once again, we rely on the notion of volume of the similarity matrix of a set of items as a measure of diversity (see Algorithm \ref{alg:marginal_gain}). If the marginal gain in volume is greater than a specified threshold $\tau$, we add the instance to the dataset ($\mathcal{S}$) and the anchor set $\Phi$  (Line 1-7). 

It is important to note that computing the marginal gain of an instance over the entire dataset (which increases in size) is computationally expensive. We approximate this by computing marginal gain with respect to instances in the anchor set, which is of a fixed size. Next, if we have rejected instances (which we track in the set $\mathcal{R}$), we know that the explorer can be refined with regard to the diversity metric. We thus perform prompt refinement using textual gradients to obtain a new set of successor explorers $\mathcal{C}$ (Lines 13-15).

\subsection{Theoretical Justification} 
Having described the main algorithm in the previous section, we now draw on connections to the mathematical theory of determinantal point processes and matrix theory to justify our algorithm. 

Our global objective is to construct a final similarity matrix $S_T$ (that corresponds to the final dataset $D_T$) with a high effective rank. As noted by ~\citet{friedman2022vendi}, who propose the Vendi Score as a diversity metric, a high effective rank of $S_T$ implies high diversity.

Let $S_T$ be an $n$ by $n$ square matrix. Let $V$ the effective rank of $S_T$ and $D$ denote the determinant of $S_T$. Let $C$ be the trace of $S$. We now state the following lemma
\begin{lemma}
The effective rank of a square matrix $S_T$ can be roughly approximated by its determinant D and the trace C, as (see Appendix \ref{lemma-proof} for proof)
\begin{equation}
    V \approx n^2 \frac{D^{1/n}}{C}, \text{with $n$ being the rows of $S_T$}
\end{equation}
\end{lemma}
Based on the above lemma, we seek to maximize $\det(S_T)$ while also noting that $\det(S_T$) is the squared volume of $D_T$. 

However, since directly maximizing $\det(S_T)$ in an iterative manner is computationally infeasible, we reduce this problem to a proxy problem -- namely, the maximum volume submatrix problem (MVS) problem. The MVS problem is to maximize the determinant of the best ($k\ \text{by}\ k$) principal sub-matrix of $S_T$ (selection of $k$ data points). 
\begin{equation}
 V_{\text{MVS}}(S_T) = \max_{\Phi \subset {D}_T, |\Phi|=k} \det(S_{\Phi})^{(1/2)}  
\end{equation}
This is a good proxy because it has been observed by ~\citet{goreinov2001maximal,cortinovis2020maximum} that the MVS is a quasi-best low-rank approximation to the original matrix $S_T$.  So we attempt to maximize the MVS of size $k$ of $S_T$ as we construct the dataset.

Unfortunately, even the above is NP-hard as we need to check all $n \choose k$ subsets of the original $S_T$. It is precisely here that we rely on the machinery of determinantal point processes to find a high-volume solution.  While determinantal point processes do not find the exact solution to the above problem, a $k$-item sample drawn from the underlying determinantal point process will favor high-volume subsets. We make use of this property of determinantal point processes as follows: (a) Define an anchor set $\Phi_{T}$ whose volume will be $\textsc{Vol}(\Phi_{T}) = (\det{S_{\Phi_{T}}})^{1/2}$. \ourmethod\ iteratively constructs this anchor set in a greedy manner, trying to maximize its volume.

More specifically, note that when we generate a batch (see Algorithm \ref{alg:explore}), we sequentially add an instance to the dataset (and to the underlying anchor set) only if the instance increases the volume of the anchor set by a specific threshold. This step seeks to make the anchor set a reservoir that is in some sense ``volume optimized'' (holding diverse points). However, at this point, we have an augmented anchor set $A_T$ whose size exceeds the fixed size ($k$). We thus prune it to size $k$ but do it once again in a manner seeking to maximize local volume. We update $\Phi_{T}$ by drawing a $k$ item sample from $A_T$ using a $k$-DPP which prefers a subset of size $k$ with high volume and will be representative of the MVS volume of $A_{T}$ and provides a ``rough'' lower bound on $V_{\text{MVS}}(S_T)$. When the algorithm finishes, we have a high-volume anchor set (by construction). This in turn translates to a high $V_{\text{MVS}}(S_T)$ which in turn translates to a high effective rank of $S_T$ and thus significantly more diverse $D_T$.

\subsection{Computational Efficiency}
Because \ourmethod\ restricts the size of the explorers and the anchor set in each iteration, our algorithm is quite computationally efficient in terms of CPU time complexity. The time complexity mainly depends on the maximum size of the anchor point set $k_{max}$, the maximum size of the candidate beam $b_{max}$, the batch size $|\mathcal{B}|$, and the maximum number of overall iterations $T$. 

Observe that the \textsc{MarginalGain} can be computed in $\mathcal{O}(k^2)$ time if the inverse of $\mathbf{K}_{sim}$ can be pre-computed and cached. This means that we can process all instances $|\mathcal{B}|$ in $\mathcal{O}(|\mathcal{B}|k^{2})$, assuming a constant time penalty for LLM calls and set addition. The \textsc{Explore} call incurs $\mathcal{O}(|\mathcal{B}|k^{2})$ cost. The pruning of the anchor set to size (Line $10$) using a DPP costs $\mathcal{O}(k_{max}^{3})$ time. Thus, the entire inner for-loop (Line 6) incurs cost  $\mathcal{O}(b(k_{max}^{3} + |\mathcal{B}|k^{2}))$. The sampling step for pruning the set of explorers using a second DPP takes $\mathcal{O}(b_{max}^{3})$ cost. The total cost incurred by the algorithm therefore is $\mathcal{O}(T(b(k_{max}^{3} + |\mathcal{B}|k^{2}) + b_{max}^{3}$)). 

Finally, note that \ourmethod\ in general is significantly more efficient than a naive approach that could use a DPP to sample the entire dataset by first generating a universe larger than the required size and then selecting from it. That would incur a cost $\mathcal{O}(|l|^{3})$ where $|l|$ is the size of the full dataset requested, which, when $|l|$ is large, can be prohibitive.

Having outlined the worst case cost of \ourmethod\, we now ask if we can roughly estimate how many iterations of the outer-loop would run in an average case. This largely depends on the fraction of data-points accepted in \textsc{Explore} step. Let $\zeta$ be the fraction of candidates accepted on an average in each call to \textsc{Explore}. That means, $\zeta|\mathcal{B}|$ instances are on average accepted per iteration. So we should expect the outer loop to run about $T_{avg}=\frac{l}{\zeta|\mathcal{B}|}$ times. Intuitively, $\zeta$ represents the ``inherent easiness'' of generating diverse data for the task using the LLM. If the LLM is easily able to generate diverse data for the task and find prompts easily to obtain diverse data, $\zeta$ will be high (and low otherwise). Finally, we note that this also gives an average estimate of the total number of LLM calls $N_{\textsc{llm}}=\frac{\rho l}{\zeta|\mathcal{B}|}$ as we make a constant number of LLM calls, $\rho$ (we have $\rho=3$) in \textsc{Explore}. 


\begin{table*}[h]
\centering
\large
\resizebox{1.0\textwidth}{!}{%
\begin{tabular}{c}
\begin{subtable}{\textwidth}
\centering
\caption{Generate a single sentence about sports}
\label{tab:sports}
\begin{tabular}{lccccc}
\toprule
\textbf{Method} & \textbf{Lexical $\uparrow$} & \textbf{Cosine $\uparrow$} & \textbf{Vendi $\uparrow$} & \textbf{Quality $\uparrow$} & \textbf{LLM  $\downarrow$} \\
\midrule
\textsc{Default} & $\num{0.670} \pm \num{0.165}$ & $\num{0.211} \pm \num{0.088}$ & $\num{2.991}$ & $\num{21.701} \pm \num{0.745}$ & 50 \\
\textsc{Temp} & $\num{0.698} \pm \num{0.151}$ & $\num{0.221} \pm \num{0.087}$ & $\num{3.234}$ & $\num{21.707} \pm \num{0.749}$ & 50 \\
\textsc{Diverse} & $\num{0.784} \pm \num{0.129}$ & $\num{0.413} \pm \num{0.174}$ & $\num{7.871}$ & $\mathbf{\num{23.104} \pm \num{0.818}}$ & 50 \\
\textsc{History} & $\num{0.794} \pm \num{0.119}$ & $\num{0.333} \pm \num{0.120}$ & $\num{5.998}$ & $\num{22.145} \pm \num{0.703}$ & 50 \\
\textsc{Hierarchical} & $\underline{\num{0.851} \pm \num{0.098}}$ & $\underline{\num{0.539} \pm \num{0.135}}$ & $\underline{\num{15.070}}$ & $\underline{\num{22.852} \pm \num{0.984}}$ & 550 \\
\textsc{SubsetSelect} & $\num{0.810} \pm \num{0.109}$ & $\num{0.414} \pm \num{0.172}$ & $\num{7.769}$ & $\num{22.223} \pm \num{0.928}$ & 500 \\
\midrule
\ourmethod & $\mathbf{\num{0.867} \pm \num{0.061}}$ & $\mathbf{\num{0.547} \pm \num{0.134}}$ & $\mathbf{\num{24.132}}$ & ${\num{22.257} \pm \num{2.060}}$ & $443$ \\
\bottomrule
\end{tabular}
\end{subtable}



\begin{subtable}{\textwidth}
\centering
\caption{Generate a short conversation about politics}
\label{tab:politics}
\begin{tabular}{lccccc}
\toprule
\textbf{Method} & \textbf{Lexical $\uparrow$} & \textbf{Cosine $\uparrow$} & \textbf{Vendi $\uparrow$} & \textbf{Quality $\uparrow$} & \textbf{LLM $\downarrow$} \\
\midrule
\textsc{Default} & $\num{0.721} \pm \num{0.053}$ & $\num{0.247} \pm \num{0.083}$ & $\num{4.589}$ & $\num{22.969} \pm \num{0.676}$ & $50$ \\
\textsc{Temp} & $\num{0.738} \pm \num{0.050}$ & $\num{0.272} \pm \num{0.086}$ & $\num{5.157}$ & $\num{23.027} \pm \num{0.673}$ & $50$ \\
\textsc{Diverse} & $\num{0.725} \pm \num{0.042}$ & $\num{0.164} \pm \num{0.035}$ & $\num{3.301}$ & $\underline{\num{23.745} \pm \num{0.881}}$ & $50$ \\
\textsc{History} & $\num{0.695} \pm \num{0.071}$ & $\num{0.262} \pm \num{0.108}$ & $\num{4.240}$ & $\num{22.966} \pm \num{0.6416}$ & $50$ \\
\textsc{Hierarchical} & $\underline{\num{0.768} \pm \num{0.052}}$ & $\underline{\num{0.382} \pm \num{0.104}}$ & $\underline{\num{8.450}}$ & $\num{23.320} \pm \num{0.616}$ & $550$ \\
\textsc{SubsetSelect} & $\num{0.728} \pm \num{0.052}$ & $\num{0.261} \pm \num{0.084}$ & $\num{4.851}$ & $\num{23.131} \pm \num{0.659}$ & $500$ \\
\midrule
\ourmethod  & $\mathbf{\num{0.820} \pm \num{0.048}}$ & $\mathbf{\num{0.439} \pm \num{0.092}}$ & $\mathbf{\num{15.035}}$ & $\mathbf{\num{23.917}\pm \num{0.882}}$ & $426$ \\
\bottomrule
\end{tabular}
\end{subtable}

\\[8em] 

\begin{subtable}{\textwidth}
\centering
\caption{Generate a poem}
\label{tab:poem}
\begin{tabular}{lccccc}
\toprule
\textbf{Method} & \textbf{Lexical $\uparrow$} & \textbf{Cosine $\uparrow$} & \textbf{Vendi $\uparrow$} & \textbf{Quality$\uparrow$} & \textbf{LLM $\downarrow$} \\
\midrule
\textsc{Default} & $\num{0.761} \pm \num{0.048}$ & $\num{0.153} \pm \num{0.046}$ & $\num{3.004}$ & $\num{22.521} \pm \num{0.889}$ & $50$ \\
\textsc{Temp} & $\num{0.779} \pm \num{0.045}$ & $\num{0.160} \pm \num{0.040}$ & $\num{3.217}$ & $\num{22.651} \pm \num{0.815}$ & $50$ \\
\textsc{Diverse} & $\num{0.781} \pm \num{0.043}$ & $\num{0.144} \pm \num{0.040}$ & $\num{2.761}$ & $\underline{\num{22.788} \pm \num{1.455}}$ & $50$ \\
\textsc{History} & $\num{0.707} \pm \num{0.054}$ & $\num{0.113} \pm \num{0.033}$ & $\num{2.295}$ & $\num{22.449} \pm \num{0.918}$ & $50$ \\
\textsc{Hierarchical} & $\underline{\num{0.820} \pm \num{0.055}}$ & $\underline{\num{0.300} \pm \num{0.080}}$ & $\underline{\num{5.679}}$ & $\num{22.561} \pm \num{1.311}$ & $550$ \\
\textsc{SubsetSelect} & $\num{0.764} \pm \num{0.048}$ & $\num{0.156} \pm \num{0.044}$ & $\num{3.082}$ & $\num{22.520} \pm \num{0.882}$ & $500$ \\

\midrule
\ourmethod  & $\mathbf{\num{0.862} \pm \num{0.047}}$ & $\mathbf{\num{0.303} \pm \num{0.084}}$ & $\mathbf{\num{7.312}}$ & $\mathbf{\num{24.505} \pm \num{0.905}}$ & $615$ \\
\bottomrule
\end{tabular}
\end{subtable}

\begin{subtable}{\textwidth}
\centering
\caption{Generate a plot for a movie}
\label{tab:movie}
\begin{tabular}{lccccc}
\toprule
\textbf{Method} & \textbf{Lexical $\uparrow$} & \textbf{Cosine $\uparrow$} & \textbf{Vendi $\uparrow$} & \textbf{Quality $\uparrow$} & \textbf{LLM $\downarrow$} \\
\midrule
\textsc{Default} & $\num{0.800} \pm \num{0.048}$ & $\num{0.207} \pm \num{0.078}$ & $\num{4.002}$ &  $\num{23.044} \pm \num{0.525}$ & $50$ \\
\textsc{Temp} & $\num{0.828} \pm \num{0.045}$ & $\num{0.261} \pm \num{0.079}$ & $\num{5.523}$ & $\num{23.077} \pm \num{0.601}$ & $50$ \\
\textsc{Diverse} & $\num{0.809} \pm \num{0.044}$ & $\num{0.254} \pm \num{0.067}$ & $\num{5.277}$ & $\mathbf{\num{23.457} \pm \num{1.994}}$ & $50$ \\
\textsc{History} & $\num{0.784} \pm \num{0.050}$ & $\num{0.219} \pm \num{0.070}$ & $\num{4.276}$ & $\underline{\num{23.149} \pm \num{0.784}}$ & $50$ \\
\textsc{Hierarchical} & $\underline{\num{0.839} \pm \num{0.047}}$ & $\underline{\num{0.322} \pm \num{0.081}}$ & $\underline{\num{7.661}}$ & $\num{22.991} \pm \num{0.924}$ & $550$ \\
\textsc{SubsetSelect} & $\num{0.809} \pm \num{0.047}$ & $\num{0.231} \pm \num{0.081}$ & $\num{4.572}$ & $\num{23.032}\pm \num{0.535}$ & $500$ \\
\midrule
\ourmethod   & $\mathbf{\num{0.840} \pm \num{0.052}}$ & $\mathbf{\num{0.341} \pm \num{0.095}}$ & $\mathbf{\num{8.302}}$ & $\num{22.959}\pm \num{1.115}$ & $695$ \\
\bottomrule
\end{tabular}
\end{subtable}
\end{tabular}
}
\caption{\small{\textbf{Creative task evaluations}. Note that \textbf{\ourmethod\ outperforms all} baselines significantly (average Vendi score improvement of \textbf{2.96} times over \textsc{Default}, $\textbf{0.43}$ times over \textsc{Hierarchical} across all creative tasks) with no significant degradation on perceived quality. \textbf{Legend} Best result in each column is highlighted in $\textbf{Bold}$ and the second-best result in each column is highlighted in \underline{underline}. Quality metrics are on a scale of $0(\text{lowest)}-25(\text{highest}$).}}
\label{tab:creative_tasks}
\end{table*}

\begin{table*}[h]
\centering
\large
\resizebox{1.0\textwidth}{!}{%
\begin{tabular}{c}
\begin{subtable}{\textwidth}
\centering
\caption{Generate a mathematical question for Grade school math}
\label{tab:math}
\begin{tabular}{lccccc}
\toprule
\textbf{Method} & \textbf{Lexical $\uparrow$} & \textbf{Cosine $\uparrow$} & \textbf{Vendi $\uparrow$} & \textbf{Quality $\uparrow$} & \textbf{LLM $\downarrow$} \\
\midrule
\textsc{Default} & $\num{0.538} \pm \num{0.130}$ & $\num{0.201} \pm \num{0.078}$ & $\num{3.039}$ & $\num{14.991} \pm \num{0.124}$ & $50$ \\
\textsc{Temp} & $\num{0.563} \pm \num{0.117}$ & $\num{0.223} \pm \num{0.076}$ & $\num{3.557}$ & $\num{14.995} \pm \num{0.097}$ & $50$ \\
\textsc{Diverse} & $\num{0.469} \pm \num{0.058}$ & $\num{0.067} \pm \num{0.024}$ & $\num{1.650}$ & $\mathbf{\num{15.000} \pm \num{0.000}}$ & $50$ \\
\textsc{History} & $\num{0.299} \pm \num{0.152}$ & $\num{0.236} \pm \num{0.116}$ & $\num{3.131}$ & $\num{14.831} \pm \num{0.579}$ & $50$ \\
\textsc{Hierarchical} & $\underline{\num{0.681} \pm \num{0.102}}$ & $\underline{\num{0.399} \pm \num{0.112}}$ & $\underline{\num{8.715}}$ & $\underline{\num{14.998} \pm \num{0.058}}$ & $550$ \\
\textsc{SubsetSelect} & $\num{0.566} \pm \num{0.117}$ & $\num{0.219} \pm \num{0.074}$ & $\num{3.477}$ & $\num{14.982} \pm \num{0.168}$ & $500$ \\
\midrule
\ourmethod  & $\mathbf{\num{0.807} \pm \num{0.056}}$ & $\mathbf{\num{0.478} \pm \num{0.096}}$ & $\mathbf{\num{18.777}}$ & $\num{14.771} \pm \num{0.919}$ & $399$ \\
\bottomrule
\end{tabular}
\end{subtable}

\begin{subtable}{\textwidth}
\centering
\caption{Generate a simple logic puzzle suitable for Grade school students}
\label{tab:puzzle}
\begin{tabular}{lccccc}
\toprule
\textbf{Method} & \textbf{Lexical $\uparrow$} & \textbf{Cosine $\uparrow$} & \textbf{Vendi $\uparrow$} & \textbf{Quality $\uparrow$} & \textbf{LLM $\downarrow$} \\
\midrule
\textsc{Default} & $\num{0.586} \pm \num{0.101}$ & $\num{0.196} \pm \num{0.089}$ & $\num{3.312}$ & $\num{14.831} \pm \num{0.692}$ & 50 \\
\textsc{Temp} & $\num{0.628} \pm \num{0.099}$ & $\num{0.252} \pm \num{0.114}$ & $\num{4.469}$ & $\num{14.798} \pm \num{0.903}$ & 50 \\
\textsc{Diverse} & $\num{0.620} \pm \num{0.077}$ & $\num{0.145} \pm \num{0.050}$ & $\num{2.825}$ & $\underline{\num{14.918} \pm \num{0.440}}$ & 50 \\
\textsc{History} & $\num{0.351} \pm \num{0.112}$ & $\num{0.265} \pm \num{0.105}$ & $\num{3.981}$ & $\mathbf{\num{14.993} \pm \num{0.132}}$ & 50 \\
\textsc{Hierarchical} & $\underline{\num{0.650} \pm \num{0.084}}$ & $\underline{\num{0.327} \pm \num{0.107}}$ & $\underline{\num{7.024}}$ & $\num{14.878} \pm \num{0.546}$ & 550 \\
\textsc{SubsetSelect} & $\num{0.624} \pm \num{0.099}$ & $\num{0.242} \pm \num{0.102}$ & $\num{4.249}$ & $\num{14.799} \pm \num{0.716}$ & 500 \\
\midrule
\ourmethod  & $\mathbf{\num{0.785} \pm \num{0.058}}$ & $\mathbf{\num{0.407} \pm \num{0.079}}$ & $\mathbf{\num{13.256}}$ & $\num{14.640} \pm \num{0.970}$ & 393 \\
\bottomrule
\end{tabular}
\end{subtable}
\end{tabular}
}
\caption{\small{\textbf{Reasoning tasks evaluations}. Note that \textbf{\ourmethod\ outperforms all} baselines significantly (average Vendi score improvement of $\textbf{4.12}$ times over \textsc{Default}, $\textbf{1.02}$ times over \textsc{Hierarchical} across all reasoning tasks)  with no significant degradation on perceived quality. \textbf{Legend} Best result in each column is highlighted in $\textbf{Bold}$ and the second-best result in each column is highlighted in \underline{underline}. Quality metrics are on a scale of $0(\text{lowest)}-15(\text{highest}$). }}
\label{tab:reasoning_tasks}
\end{table*}

\section{Experiments}
In this section, we report on comprehensive experiments to evaluate our proposed method on its effectiveness to generate diverse datasets. 
\subsection{Generation Tasks and Evaluation Metrics}
\paragraph{Generation Tasks} We consider the two different categories of text generation tasks: (a) creative writing and (b) reasoning to evaluate our method against strong baselines. In the creative writing category, we consider four generation tasks: (a) topical sentence generation, (b) conversation generation (c) poem generation, and (d) movie plot generation. In the reasoning category, we consider the tasks of (a)  grade school mathematical question generation and (b) Logical reasoning questions.

\paragraph{Diversity Metrics} Because there is no universal consensus on metrics for evaluating diversity, we consider a few popular metrics for diversity capturing different facets.

\begin{itemize}[noitemsep]
\item \textbf{Cosine Distance}: We consider the mean cosine distance between pairs of instances as a measure of diversity.


\item \textbf{Lexical Distance}: While the cosine distance metric captures semantic diversity, it may not capture lexical diversity very well. Therefore, we also consider a lexical diversity metric. To compute lexical diversity, we use the Jaccard distance and report the mean Jaccard distance (removing stop words) over pairs of instances.  
\item \textbf{Vendi Score}: Vendi score \cite{friedman2022vendi} is a popular diversity metric that seeks to capture diversity of the dataset in a broader and general sense (potentially including all the above dimensions). It seeks to measure the effective number of data instances that can represent the data. 
\end{itemize}

\paragraph{Quality Metric} We also seek to capture the impact on quality of the generations as we seek to encourage diversity. Consequently, we also measure the quality of the generations using an LLM-as-Judge framework where the judge evaluates the generations on a task-specific rubric (eg, dimensions like faithfulness, coherence, etc). 

\paragraph{LLM Calls} We also track the number of LLM calls each method to generate the same size dataset (assuming all methods call the LLM to generate the same batch size of data).

\subsection{Baseline Methods} We consider the following baselines:
\begin{itemize}[noitemsep]
\item \textbf{Vanilla Generation - \textsc{Default}}: This simply prompts the LLM (with default parameters) to perform the generation task.

\item \textbf{Temperature based sampling - \textsc{Temp}}: One lever to encourage diversity in LLM generations is the temperature parameter used to scale the LLM output logits when sampling tokens. We set higher temperature to $2.0$.

\item \textbf{Generate ``diverse'' command - \textsc{Diverse}}: This explicitly ask the LLM to be "diverse" in its output by appending it to the user task instructions appropriately.

\item \textbf{Conditioned on history -- \textsc{History}}: Another approach to encourage diversity is to provide a history of prior generated data instances (of a fixed window) and ask it to avoid generating such instances.

\item \textbf{Hierarchical Prompting - \textsc{Hierarchical}}: Here we ask the LLM to first generate K diverse topics and then actual instances conditioned on those topics to encourage diversity.

\item \textbf{Subset Select - \textsc{SubsetSelect}}:  We evaluate whether just using a $k$-DPP to sample the dataset from a larger set (universe) generated using Vanilla sampling would be effective. 



\end{itemize}

\subsection{Experimental Settings}
\paragraph{Similarity Kernel Functions} Our similarity kernel function is a convex combination of a radial basis function kernel (RBF) using embeddings of the text, and a lexical similarity kernel using Jaccard similarity. RBF kernel is well known to be positive semi-definite (PSD), and Jaccard Similarity has also been proved to be PSD \cite{article}. A convex combination of two PSD kernels maintains the PSD property. The weights of the convex combination are $0.7$ and $0.3$ for the radial basis function kernel and the lexical similarity kernel, respectively.

\paragraph{LLMs} All experimental data were generated using GPT-4o mini. For text embeddings, we utilize OpenAI's text-embedding-3-small model.

\paragraph{\textsc{Hierarchical} baseline settings} For the hierarchical baseline, in each call, we generate $10$ subtopics to ensure comprehensive coverage of the problem space. We then generate $1$ sample for each such topic to return $10$ instances.

\paragraph{\ourmethod\ Hyperparameters} Our experiment settings are as follows: \textbf{Number of explorers ($b$)}: $3$, 
\textbf{Number of anchor points ($k$)}: $10$, 
\textbf{Maximum iterations ($T$)}: $200$, 
\textbf{Samples per LLM call ($|\mathcal{B}|$)}: $10$, 
\textbf{Target dataset size ($l$)}: $500$. For initializing $\tau$, we followed the initialization procedure in Appendix \ref{set_tau}. See Appendix \ref{refinement-prompts} for the meta-prompts used in the refinement procedure.

\paragraph{Evaluation Jury} All generated outputs were evaluated by a panel consisting of GPT-4, GPT-4o, and GPT-4.1, providing a robust quality assessment framework. We use the mean function to obtain jury consensus scores for each dimension in the rubric. We use the standard rubrics as in \cite{paech2024eqbenchemotionalintelligencebenchmark} for poem generation. For other tasks, we use this template and modify it to be more task-specific. The prompts used for jury evaluation are available in the Appendix \ref{quality-prompts}.

\subsection{Results and Discussion}
\paragraph{Creating Writing Tasks} Tables \ref{tab:creative_tasks} show the results for the creative writing tasks, from which we can make the following observations. 

First, note that temperature based methods improve over the baseline in terms of all diversity metrics with no significant decrease in quality. 

Second, with regards to the ``prompt-diversity'' control method, the \textsc{Diverse-Keyword} baseline significantly improves diversity (across all metrics) over both the default baseline and temperature based baselines, suggesting that explicit instructions to make the output diverse is helpful. 

Third, the history based prompting outperforms the baseline marginally but not as strongly as the explicit \textsc{Diverse Keyword} approach. This may be because the \textsc{History} based approach seeks to ensure non-redundancy with the prior history (of a fixed size) but that does not necessarily yield diverse outputs overall. One advantage of these baselines is that to construct a dataset of size $l$ with a batch size $|\mathcal{B}|$ they will make exactly $\frac{l}{|\mathcal{B}|}$ LLM calls irrespective of the task. However, this efficiency in terms of LLM calls comes at a significant cost (significant loss of diversity). \ourmethod\ in contrast enforces quality control to obtain diversity and thus generally pays a cost in terms of more LLM calls. In practice, we opine that the right choice in the tradeoff of LLM-Calls vs Diversity gains is task and application-specific. For example, one clear application is around user persona modeling, where we may want to synthesize data for user personalization tasks (simulate user personas), diversity is critical for the success. In this case the benefit far exceeds the cost.

Fourth, the hierarchical prompting approach very significantly outperforms other methods in this class. This suggests that incorporating domain knowledge and explicit instructions to explore different topics significantly improves diversity. 

Finally, all flavors of \ourmethod\ significantly outperform all other methods, significantly suggesting the effectiveness of \ourmethod\ as we directly optimize a quantitative measure of diversity (volume) with no significant degradation in quality. It is also worth noting that \ourmethod\ also makes significantly fewer LLM calls (in most cases) compared to \textsc{Hierarchy}, the best competing baseline, and suggests that \ourmethod\ has a generally better cost-benefit tradeoff (LLM calls vs diversity).

\begin{figure}[t] 
    \centering 
    \includegraphics[trim={0 1cm 0 2cm },clip, width=\columnwidth]{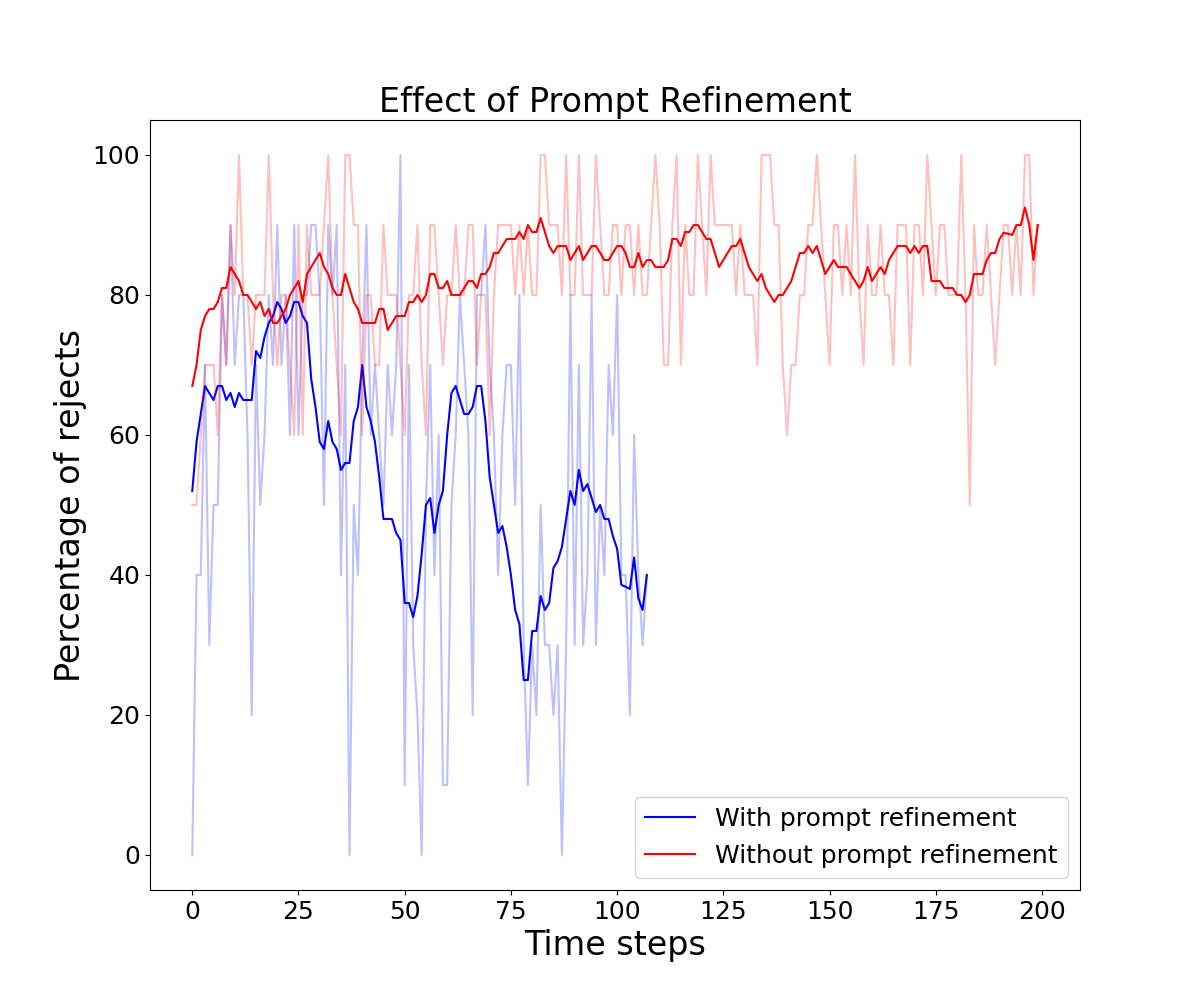}
    \caption{\textbf{Rejection rate of samples} within a batch over time with ``textual gradients'' enabled vs disabled to generate the same dataset size for the sports task (all other settings identical). }
    \label{fig:abalation_prompt} 
\end{figure}

\paragraph{Reasoning Tasks} Tables \ref{tab:reasoning_tasks} show the results of the experiments for the reasoning tasks. Many of the observations noted in the creative writing experiments hold for the reasoning tasks as well. Most notably, \ourmethod\ significantly outperforms all baselines on the reasoning tasks as well. 

\begin{table}[htpb] 
\centering
\resizebox{\linewidth}{!}{%
\begin{tabular}{l|cc|cc}
\toprule
\textbf{Task} & \textbf{DEFAULT} & \textbf{\ourmethod} & \textbf{FLEISS KAPPA} & \textbf{KRIPPENDORFF $\alpha$}\\
\midrule
Sports & $2.16 \pm 0.55$ & $\mathbf{3.82 \pm 0.28}$ & $0.41$ & $0.74$\\
Math & $1.56 \pm 0.36$ & $\mathbf{3.72 \pm 0.33}$ & $0.34$ & $0.72$\\
\bottomrule
\end{tabular}
}
\caption{\small{\textbf{Human-rated diversity}. Rubrics: 1 = Very Similar, 2=Similar, 3 = Dissimilar, 4 = Very Dissimilar.  Note that \textsc{\ourmethod} has a higher mean (lower variance) diversity compared to the \textsc{DEFAULT} baseline in both the Sports and Math tasks.}}
\label{tab:human-table-main}
\end{table}
\paragraph{Human Evaluations \& Qualitative results} 
To complement our automatic diversity scores (Vendi score), we conducted a human evaluation on Sports and Math tasks. Each task contains 200 pairs of samples, consisting of 100 pairs from the DEFAULT method and 100 pairs from the $\ourmethod$ method. The source of each pair is intentionally masked to ensure unbiased evaluation; 3 annotators rated the diversity using the following evaluation rubrics.

Table \ref{tab:human-table-main} shows the human-rated diversity (3 annotators) of \textsc{\ourmethod} and \textsc{default} baseline for Sports and Math tasks. Note that \textsc{\ourmethod} achieves a higher mean diversity score compared to the \textsc{default} baseline, indicating the correlation with automatic diversity metrics. 

\textbf{Evaluation Rubric:} Please compare both the samples in each pair and assign a score from 1 to 4 based on their similarity: 1 = Very Similar, 2= Similar, 3 = Dissimilar, 4 = Very Dissimilar 

\textbf{Inter annotator agreement: }
Table \ref{tab:human-table-main} also shows inter-annotator agreement (Fleiss' Kappa, Krippendorff's alpha), which reveal that there is good level of inter-annotator reliability.


See Appendix \ref{quali-tab} for qualitative results on  Sports and Math tasks.


\subsection{Ablation Studies}
\paragraph{Effect of Diverse Explorers} To evaluate the impact of encouraging diversity in the explorers by sampling from a DPP, we replace the selection of successors to just be a random sample of the same size since it is not critical to our algorithm. We call this algorithm variant \ourmethod-\textsc{RE}(RandomExplorer). The main advantage of \ourmethod-\textsc{RE} is that we do not need to incur the cost of sampling from a $k$-DPP, which $\O(|\mathcal{C}|^{3})$ in each main iteration of the loop. This potentially trades off some diversity and efficient convergence for reduced local computational time (a choice that practitioners have), which we validate empirically in Table \ref{tab:abalation2}. Note that \ourmethod-\textsc{RE} shows a drop in diversity,  quality and requires more LLM calls to generate the same number of instances as \ourmethod\ confirming our observation above.


\begin{table}[t] 
\centering
\resizebox{\linewidth}{!}{
\begin{tabular}{lccccc}
\toprule
\textbf{Method} & \textbf{Lexical $\uparrow$} & \textbf{Cosine $\uparrow$}  & \textbf{Vendi $\uparrow$} & \textbf{Quality  $\uparrow$} & \textbf{LLM $\downarrow$} \\
\midrule
\ourmethod-\textsc{RE}  & ${\num{0.851} \pm \num{0.072}}$ & ${\num{0.443} \pm \num{0.144}}$ & $\num{11.852}$ & $\num{21.473} \pm \num{1.881}$ & 361 \\
\ourmethod  & $\mathbf{\num{0.854} \pm \num{0.072}}$ & $\mathbf{\num{0.472} \pm \num{0.152}}$ & $\mathbf{\num{14.282}}$ & $\mathbf{\num{21.875} \pm \num{1.513}}$ & $\mathbf{252}$ \\
\bottomrule
\end{tabular}
}
\caption{\small{\textbf{Ablation for DPP} used in selecting $b$ explorers on the sports task. Observe that choosing a random sample of explorers results in slightly lower diversity and also incurs more LLM calls as opposed to choosing a diverse set of explorers confirming our hypothesis that diverse explorers can make search through the space more effective.}}
\label{tab:abalation2}
\end{table}

\noindent \textbf{Effect of Textual Gradients} We also ablate the prompt refinement step using textual gradients to evaluate its effectiveness. In particular, while we still attempt to maximize our operationalize notion of diversity by selecting only instances which a marginal gain above a threshold, we disable prompt refinement. This means no new explorers will be generated and we will only use the same explorer many times (with variance in output only due to the LLM call). We compare this with a setting (where prompt refinement is enabled) but restrict the beam size to exactly be 1, so that we have an ``apples-to-apples'' comparison with the setting where the same explorer is used across all iterations. Our hypothesis is that disabling the prompt refinement would result in a significantly higher average rejection rate (the fraction of instances rejected because they did not meet the marginal gain criteria in \textsc{Explore} step), and the algorithm would have to run for many more iterations. Figure \ref{fig:abalation_prompt} shows the results of this ablation. Observe that disabling the textual gradients results in a much higher average rejection rate and number of iterations, confirming our hypothesis that allowing for feedback to refine the prompt can help make the search more efficient.

\noindent \textbf{Effect of Generation Length on Diversity}
While all methods have the same setting for maximum output length of $2048$ tokens, qualitatively we observed that baseline methods tend to generate shorter length generations (of their own accord) while our method recognizes that length can be an important lever to generate diverse data and leverages it smartly. Still, one might ask whether most of \ourmethod's diversity gains can be attributed to its responses being significantly longer. To get at this, we conduct another experiment where we explicitly prompt LLMs (all methods including ours) to generate about the same number of tokens. This experiment thus controls for output length (as much as possible). As shown in the Table \ref{tab:poem-lenght}, we note in this case as well that our method (Voyager) outperforms other baseline methods, suggesting that our method’s gains on diversity do not stem from length alone (although we believe length is an important aspect of diversity).

\begin{table}[t]
\centering
\caption{\textbf{Ablation study to control for generation length} The  task prompt is ``Generate a poem. Restrict the generated output to 150 tokens". \ourmethod\  outperforms other methods on diversity even when the average token counts for all methods are very similar.}
\resizebox{\linewidth}{!}{%
\label{tab:poem-lenght}
\begin{tabular}{lcccccc}
\toprule
\textbf{Method} & \textbf{Lexical $\uparrow$} & \textbf{Cosine $\uparrow$} & \textbf{Vendi $\uparrow$} & \textbf{Quality$\uparrow$} & \textbf{LLM $\downarrow$} &  \textbf{Token count} \\
\midrule
\textsc{Default} & $\num{0.775} \pm \num{0.053}$ & $\num{0.169} \pm \num{0.042}$ & $\num{3.304}$ & $\num{22.421} \pm \num{0.924}$ & $50$ & $\num{131.28} \pm \num{12.761}$ \\
\textsc{Temp} & $\num{0.784} \pm \num{0.050}$ & $\num{0.171} \pm \num{0.040}$ & $\num{3.432}$ & $\num{22.499} \pm \num{0.874}$ & $50$ & $\num{132.676} \pm \num{12.891}$ \\
\textsc{Diverse} & $\num{0.778} \pm \num{0.050}$ & $\num{0.163} \pm \num{0.035}$ & $\num{3.263}$ & $\num{22.452} \pm \num{0.891}$ & $50$ & $\num{132.792} \pm \num{13.344}$ \\
\textsc{History} & $\num{0.754} \pm \num{0.053}$ & $\num{0.160} \pm \num{0.046}$ & $\num{3.095}$ & $\num{22.405} \pm \num{0.962}$ & $50$ & $\num{143.66} \pm \num{13.345}$ \\
\textsc{Hierarchical} & $\num{0.837} \pm \num{0.056}$ & $\num{0.313} \pm \num{0.077}$ & $\num{6.646}$ & $\num{22.153} \pm \num{1.325}$ & $550$ & $\num{136.152} \pm \num{13.270}$ \\
\textsc{SubsetSelect} & $\num{0.778} \pm \num{0.052}$ & $\num{0.173} \pm \num{0.042}$ & $\num{3.399}$ & $\num{22.417} \pm \num{0.928}$ & $500$ & $\num{132.52} \pm \num{12.943}$ \\
\ourmethod  & $\mathbf{\num{0.859} \pm \num{0.0485} } $ & $\mathbf{\num{0.330} \pm \num{0.0863}}  $ & $\mathbf{\num{8.167} } $ & $\mathbf{\num{23.583} \pm \num{1.2204}}  $ & $640$ & $\num{142.398} \pm \num{13.975} $ \\
\bottomrule
\end{tabular}
}
\end{table}


\subsection{Effectiveness of Data for Synthetic Training Data Generation}
 Finally, we also evaluate whether a model trained on data using \ourmethod\ would result in higher model performance compared to a model trained on synthetic data generated by the \textsc{Default} baseline.

We consider the GSM8K task and generate $1000$ questions using the prompt ``Generate a mathematical question for Grade School Math''. We use 5-shot prompting for GPT4 \cite{openai2024gpt4technicalreport} to answer these questions to obtain (question, answer) pairs for the training data. We then train Gemma-2b-it and Gemma-7b-it models (instruction-tune) using just these $1000$ examples (refer Appendix \ref{gsm9k-train} for training details). We compare against models trained on the same number of examples but generated using a \textsc{Default} baseline. We evaluate both settings on the standard GSM8K test set \cite{cobbe2021trainingverifierssolvemath}. Results are shown in Table \ref{tab:predictive}. We note that data generated using \ourmethod\ yields a significantly higher model performance than the baseline, suggesting the effectiveness of our method and more generally the effectiveness of training on diverse data ($35.7$ vs $45.7$ -- see column Gemma-7b-it). In fact, we noted that using only $500$ examples, we were able to almost match (and even outperform) the performance of models trained on data generated by the \textsc{Default} baseline, further underscoring the importance of diversity in training data, an observation noted by several prior works \cite{bukharin2024data,pang2024improving}

\begin{table}[t]
\centering
\resizebox{\linewidth}{!}{%
\begin{tabular}{lcccc}
\toprule
\textbf{Method} & $|D_{train}|$ & \textbf{LLM $\downarrow$} & \textbf{ Gemma 2B-IT} & \textbf{Gemma 7B-IT }\\
\midrule
\textsc{Default} & $1000$ & $100$ & \underline{13.1} & 35.7\\
\ourmethod  & $1000$ & $793$ &$\mathbf{16.4}$ & $\mathbf{45.7}$ \\
\ourmethod & $500$ & $399$ &12.9 & \underline{42.8} \\
\bottomrule
\end{tabular}
}
\caption{\small{\textbf{GSM8K zero-shot test accuracy on Gemma} trained on data generated using \ourmethod\  compared to the \textsc{Default} baseline. Note that \ourmethod\ which generates more diverse data leads to improved predictive performance and training data efficiency compared to the baseline.}}
\label{tab:predictive}
\end{table}

\section{Conclusion}
We proposed \ourmethod, a training free approach to use LLMs to generate more diverse data. Our method relies on the determinant of the similarity matrix of a dataset is related to the volume spanned by the data to measure data diversity. Building on this observation, we propose an iterative algorithm that seeks to approximately maximize this measure of diversity by leveraging the machinery of determinantal point processes and prompt refinement using textual gradients. To conclude, our method significantly improves the diversity of generated data by $1.5-3$x, is training-free and scalable, compared to baseline methods.   

\section*{Limitations}
Our work is not without its limitations. First, \ourmethod\ depends on high-quality instruction following LLMs. This is necessary during sample generation, but more importantly, during the ``textual gradient extraction" phase, where the LLM must analyze a prompt, its generation, and then provide suggestions for improvement. Secondly, the similarity computations require a robust embedding model that can effectively differentiate differences in text within the embedding space. Furthermore, a comprehensive analysis of the (type of) diversity introduced by \ourmethod\ is essential to fully understand its impact and effectiveness. Finally, this work focuses solely on diversity in text generation using LLMs. We do not consider diversity of multi-modal data, which poses additional challenges, such as how to measure similarity of representations across multiple modalities.

\section*{Acknowledgments}

We would like to express our gratitude to Abhinav Kumar and the anonymous reviewers for their constructive feedback, which significantly contributed to the improvement of this work.




\bibliography{custom}

@misc{pryzant2023automaticpromptoptimizationgradient,
      title={Automatic Prompt Optimization with "Gradient Descent" and Beam Search}, 
      author={Reid Pryzant and Dan Iter and Jerry Li and Yin Tat Lee and Chenguang Zhu and Michael Zeng},
      year={2023},
      eprint={2305.03495},
      archivePrefix={arXiv},
      primaryClass={cs.CL},
      url={https://arxiv.org/abs/2305.03495}, 
}

@article{kulesza2012determinantal,
  title={Determinantal point processes for machine learning},
  author={Kulesza, Alex and Taskar, Ben and others},
  journal={Foundations and Trends{\textregistered} in Machine Learning},
  volume={5},
  number={2--3},
  pages={123--286},
  year={2012},
  publisher={Now Publishers, Inc.}
}

@misc{paech2024eqbenchemotionalintelligencebenchmark,
      title={EQ-Bench: An Emotional Intelligence Benchmark for Large Language Models}, 
      author={Samuel J. Paech},
      year={2024},
      eprint={2312.06281},
      archivePrefix={arXiv},
      primaryClass={cs.CL},
      url={https://arxiv.org/abs/2312.06281}, 
}

@article{article,
author = {Bouchard, Mathieu and Jousselme, Anne-Laure and Doré, Pierre-Emmanuel},
year = {2013},
month = {07},
pages = {615 - 626},
title = {A proof for the positive definiteness of the Jaccard index matrix},
volume = {54},
journal = {International Journal of Approximate Reasoning},
doi = {10.1016/j.ijar.2013.01.006}
}

@misc{holtzman2020curiouscaseneuraltext,
      title={The Curious Case of Neural Text Degeneration}, 
      author={Ari Holtzman and Jan Buys and Li Du and Maxwell Forbes and Yejin Choi},
      year={2020},
      eprint={1904.09751},
      archivePrefix={arXiv},
      primaryClass={cs.CL},
      url={https://arxiv.org/abs/1904.09751}, 
}

@misc{jiang2025artificialhivemindopenendedhomogeneity,
      title={Artificial Hivemind: The Open-Ended Homogeneity of Language Models (and Beyond)}, 
      author={Liwei Jiang and Yuanjun Chai and Margaret Li and Mickel Liu and Raymond Fok and Nouha Dziri and Yulia Tsvetkov and Maarten Sap and Alon Albalak and Yejin Choi},
      year={2025},
      eprint={2510.22954},
      archivePrefix={arXiv},
      primaryClass={cs.CL},
      url={https://arxiv.org/abs/2510.22954}, 
}

@InProceedings{pmlr-v97-kool19a,
  title = 	 {Stochastic Beams and Where To Find Them: The {G}umbel-Top-k Trick for Sampling Sequences Without Replacement},
  author =       {Kool, Wouter and Van Hoof, Herke and Welling, Max},
  booktitle = 	 {Proceedings of the 36th International Conference on Machine Learning},
  pages = 	 {3499--3508},
  year = 	 {2019},
  editor = 	 {Chaudhuri, Kamalika and Salakhutdinov, Ruslan},
  volume = 	 {97},
  series = 	 {Proceedings of Machine Learning Research},
  month = 	 {09--15 Jun},
  publisher =    {PMLR},
  url = 	 {https://proceedings.mlr.press/v97/kool19a.html}
}

@misc{cobbe2021trainingverifierssolvemath,
      title={Training Verifiers to Solve Math Word Problems}, 
      author={Karl Cobbe and Vineet Kosaraju and Mohammad Bavarian and Mark Chen and Heewoo Jun and Lukasz Kaiser and Matthias Plappert and Jerry Tworek and Jacob Hilton and Reiichiro Nakano and Christopher Hesse and John Schulman},
      year={2021},
      eprint={2110.14168},
      archivePrefix={arXiv},
      primaryClass={cs.LG},
      url={https://arxiv.org/abs/2110.14168}, 
}

@misc{openai2024gpt4technicalreport,
      title={GPT-4 Technical Report}, 
      author={OpenAI and Josh Achiam and Steven Adler and Sandhini Agarwal and Lama Ahmad and others.},
      year={2024},
      eprint={2303.08774},
      archivePrefix={arXiv},
      primaryClass={cs.CL},
      url={https://arxiv.org/abs/2303.08774}, 
}

@article{friedman2022vendi,
  title={The vendi score: A diversity evaluation metric for machine learning},
  author={Friedman, Dan and Dieng, Adji Bousso},
  journal={arXiv preprint arXiv:2210.02410},
  year={2022}
}

@article{goreinov2001maximal,
  title={The maximal-volume concept in approximation by low-rank matrices},
  author={Goreinov, Sergei A and Tyrtyshnikov, Eugene E},
  journal={Contemporary Mathematics},
  volume={280},
  pages={47--52},
  year={2001},
  publisher={Providence, RI; American Mathematical Society; 1999}
}

@article{cortinovis2020maximum,
  title={On maximum volume submatrices and cross approximation for symmetric semidefinite and diagonally dominant matrices},
  author={Cortinovis, Alice and Kressner, Daniel and Massei, Stefano},
  journal={Linear Algebra and its Applications},
  volume={593},
  pages={251--268},
  year={2020},
  publisher={Elsevier}
}

@article{wright2025epistemic,
  title={Epistemic Diversity and Knowledge Collapse in Large Language Models},
  author={Wright, Dustin and Masud, Sarah and Moore, Jared and Yadav, Srishti and Antoniak, Maria and Park, Chan Young and Augenstein, Isabelle},
  journal={arXiv preprint arXiv:2510.04226},
  year={2025}
}

@article{padmakumar2023does,
  title={Does writing with language models reduce content diversity?},
  author={Padmakumar, Vishakh and He, He},
  journal={arXiv preprint arXiv:2309.05196},
  year={2023}
}

@inproceedings{bukharin2024data,
  title={Data diversity matters for robust instruction tuning},
  author={Bukharin, Alexander and Li, Shiyang and Wang, Zhengyang and Yang, Jingfeng and Yin, Bing and Li, Xian and Zhang, Chao and Zhao, Tuo and Jiang, Haoming},
  booktitle={Findings of the Association for Computational Linguistics: EMNLP 2024},
  pages={3411--3425},
  year={2024}
}

@inproceedings{
amballa2025quasirandom,
title={Quasi-random Multi-Sample Inference for Large Language Models},
author={Avinash Amballa and Aditya Parashar and Aditya Vikram Singh and Jinlin Lai and Benjamin Rozonoyer},
booktitle={Frontiers in Probabilistic Inference: Learning meets Sampling},
year={2025},
url={https://openreview.net/forum?id=aUAExbvozj}
}

@article{samvelyan2024rainbow,
  title={Rainbow teaming: Open-ended generation of diverse adversarial prompts},
  author={Samvelyan, Mikayel and Raparthy, Sharath C and Lupu, Andrei and Hambro, Eric and Markosyan, Aram H and Bhatt, Manish and Mao, Yuning and Jiang, Minqi and Parker-Holder, Jack and Foerster, Jakob and others},
  journal={Advances in Neural Information Processing Systems},
  volume={37},
  pages={69747--69786},
  year={2024}
}

@article{pang2024improving,
  title={Improving data efficiency via curating llm-driven rating systems},
  author={Pang, Jinlong and Wei, Jiaheng and Shah, Ankit Parag and Zhu, Zhaowei and Wang, Yaxuan and Qian, Chen and Liu, Yang and Bao, Yujia and Wei, Wei},
  journal={arXiv preprint arXiv:2410.10877},
  year={2024}
}

@article{chen2025post,
  title={Post-training Large Language Models for Diverse High-Quality Responses},
  author={Chen, Yilei and Chakraborty, Souradip and Wolf, Lorenz and Paschalidis, Yannis and Pacchiano, Aldo},
  journal={arXiv preprint arXiv:2509.04784},
  year={2025}
}

@article{li2025jointly,
  title={Jointly reinforcing diversity and quality in language model generations},
  author={Li, Tianjian and Zhang, Yiming and Yu, Ping and Saha, Swarnadeep and Khashabi, Daniel and Weston, Jason and Lanchantin, Jack and Wang, Tianlu},
  journal={arXiv preprint arXiv:2509.02534},
  year={2025}
}

@article{yao2025diversity,
  title={Diversity-Aware Policy Optimization for Large Language Model Reasoning},
  author={Yao, Jian and Cheng, Ran and Wu, Xingyu and Wu, Jibin and Tan, Kay Chen},
  journal={arXiv preprint arXiv:2505.23433},
  year={2025}
}

@article{vijayakumar2016diverse,
  title={Diverse beam search: Decoding diverse solutions from neural sequence models},
  author={Vijayakumar, Ashwin K and Cogswell, Michael and Selvaraju, Ramprasath R and Sun, Qing and Lee, Stefan and Crandall, David and Batra, Dhruv},
  journal={arXiv preprint arXiv:1610.02424},
  year={2016}
}

@article{wang2024diversity,
  title={Diversity measurement and subset selection for instruction tuning datasets},
  author={Wang, Peiqi and Shen, Yikang and Guo, Zhen and Stallone, Matthew and Kim, Yoon and Golland, Polina and Panda, Rameswar},
  journal={arXiv preprint arXiv:2402.02318},
  year={2024}
}

@article{ismayilzada2025creative,
  title={Creative preference optimization},
  author={Ismayilzada, Mete and Laverghetta Jr, Antonio and Luchini, Simone A and Patel, Reet and Bosselut, Antoine and van der Plas, Lonneke and Beaty, Roger},
  journal={arXiv preprint arXiv:2505.14442},
  year={2025}
}

@article{lanchantin2025diverse,
  title={Diverse preference optimization},
  author={Lanchantin, Jack and Chen, Angelica and Dhuliawala, Shehzaad and Yu, Ping and Weston, Jason and Sukhbaatar, Sainbayar and Kulikov, Ilia},
  journal={arXiv preprint arXiv:2501.18101},
  year={2025}
}

@inproceedings{hewitt2022truncation,
  title={Truncation Sampling as Language Model Desmoothing},
  author={Hewitt, John and Manning, Christopher D and Liang, Percy},
  booktitle={Findings of the Association for Computational Linguistics: EMNLP 2022},
  pages={3414--3427},
  year={2022}
}

@inproceedings{minhturning,
  title={Turning Up the Heat: Min-p Sampling for Creative and Coherent LLM Outputs},
  author={Minh, Nguyen Nhat and Baker, Andrew and Neo, Clement and Roush, Allen G and Kirsch, Andreas and Shwartz-Ziv, Ravid},
  booktitle={The Thirteenth International Conference on Learning Representations}
}

@article{chang2024real,
  title={REAL Sampling: Boosting Factuality and Diversity of Open-Ended Generation via Asymptotic Entropy},
  author={Chang, Haw-Shiuan and Peng, Nanyun and Bansal, Mohit and Ramakrishna, Anil and Chung, Tagyoung},
  journal={arXiv preprint arXiv:2406.07735},
  year={2024}
}

@INPROCEEDINGS{olivier2007effective,
  author={Roy, Olivier and Vetterli, Martin},
  booktitle={2007 15th European Signal Processing Conference}, 
  title={The effective rank: A measure of effective dimensionality}, 
  year={2007},
  volume={},
  number={},
  pages={606-610},
  doi={}}

@article{mehrotra2024enhancing,
  title={Enhancing creativity in large language models through associative thinking strategies},
  author={Mehrotra, Pronita and Parab, Aishni and Gulwani, Sumit},
  journal={arXiv preprint arXiv:2405.06715},
  year={2024}
}

@inproceedings{tian2024macgyver,
  title={MacGyver: Are Large Language Models Creative Problem Solvers?},
  author={Tian, Yufei and Ravichander, Abhilasha and Qin, Lianhui and Le Bras, Ronan and Marjieh, Raja and Peng, Nanyun and Choi, Yejin and Griffiths, Thomas L and Brahman, Faeze},
  booktitle={Proceedings of the 2024 Conference of the North American Chapter of the Association for Computational Linguistics: Human Language Technologies (Volume 1: Long Papers)},
  pages={5303--5324},
  year={2024}
}

@inproceedings{lu2025benchmarking,
  title={Benchmarking language model creativity: A case study on code generation},
  author={Lu, Yining and Wang, Dixuan and Li, Tianjian and Jiang, Dongwei and Khudanpur, Sanjeev and Jiang, Meng and Khashabi, Daniel},
  booktitle={Proceedings of the 2025 Conference of the Nations of the Americas Chapter of the Association for Computational Linguistics: Human Language Technologies (Volume 1: Long Papers)},
  pages={2776--2794},
  year={2025}
}

@article{wong2024simplestrat,
  title={Simplestrat: Diversifying language model generation with stratification},
  author={Wong, Justin and Orlovskiy, Yury and Luo, Michael and Seshia, Sanjit A and Gonzalez, Joseph E},
  journal={arXiv preprint arXiv:2410.09038},
  year={2024}
}

\appendix

\newpage
\clearpage

\section{Appendix}
\label{sec:appendix}

\subsection{\ourmethod - Algorithm}

\begin{algorithm}[htpb]
\caption{$\textsc{ComputeMarginalGain}(\cdot)$}
\label{alg:marginal_gain}
\begin{algorithmic}[3]
\REQUIRE $w$: Item whose marginal gain needs to be computed \newline
$\mathbf{K}$: Similarity kernel function to use.\newline
$A$: Set of items already present 
\ENSURE $\gamma$: Marginal gain in volume if $\mathbf{w}$ is added to the set.
\STATE  $\mathbf{S} = \mathbf{K}(A)$ \algorithmiccomment{Construct Similarity kernel}
\STATE  $\mathbf{S'} = \mathbf{K}(A\cup w)$
\STATE  $\gamma = \frac{\det(\mathbf{S'})}{\det(\mathbf{S})}$
\STATE return $\gamma$
\end{algorithmic}
\end{algorithm}

\begin{algorithm}[htpb]
\caption{$\textsc{SampleDPP}(\cdot)$}
\label{alg:SampleDPP}
\begin{algorithmic}[3]
\REQUIRE $U$: Set of items \newline
$n$: number of items to sample \newline
$\mathbf{K}$: Similarity kernel function to use.
\ENSURE $\mathbf{S}$: A subset of $n$ items sampled from underlying DPP.
\STATE  $\mathbf{L} = \mathbf{K}(U)$  \algorithmiccomment{Construct likelihood kernel}
\STATE $\mathbf{S}$ = Sample \textsc{K-DPP}($\mathbf{L}$, n)  \cite{kulesza2012determinantal}
\STATE return $\mathbf{S}$
\end{algorithmic}
\end{algorithm}

\subsection{Recommendation for setting $\tau$}
\label{set_tau}
\begin{algorithm}[htpb]
\caption{Recommended Initialization of $\tau$ }
\label{alg:DPP}
\begin{algorithmic}[3]
\REQUIRE $M$: LLM \\
$p$: task prompt \\
$\mathbf{K}$: similarity kernel
\ENSURE $\tau_0$: Initial threshold
\STATE Sample $\mathbf{Y} = y_1, y_2, \dots y_{100} \sim  M(p)$
    \STATE $\mathbf{S} \leftarrow \textsc{SampleDPP}(\mathbf{Y}, n=10, \mathbf{K})$
    \STATE $\mathbf{K}_S = \mathbf{K}(\mathbf{S})$ \algorithmiccomment{Construct Similarity kernel}
    \STATE $\tau_0 = \textsc{Clip} (\alpha \cdot \det(\mathbf{K}_S), \tau_{min}, \tau_{max})$
    \STATE return $\tau_0$
\end{algorithmic}
\end{algorithm}

As an expedient choice we recommend decaying $\tau = \tau_0 * \exp (-\frac{i}{T})$ for faster convergence. Note that we did not invest effort in finetuning the parameters like $\tau$ of our method.

\subsection{Proof of Lemma 1} \label{lemma-proof}
\begin{proof}
To recap, $S_T$ is a square $n$ by $n$ similarity matrix, $V$ denotes the effective rank of $S_T$, and $D$ denote the determinant of $S_T$. Let $C$ be the trace of $S_T$. Let $\lambda_{1}, \lambda_{2},\cdot, \cdot, \cdot,\lambda_{n}$ be the eigenvalues of $S_T$. Let $p_i \triangleq \frac{\lambda_{i}}{C}$. $p_i$ is just the normalized eigenvalue so that the set of $p_i$ forms a probability distribution.


We use Taylor approximations of two mathematical quantities: (a) Shannon Entropy of normalized eigen values defined as $H = -\sum p_{i} \ln p_{i}$ and (b) Weiner entropy of the normalized eigen values defined as $W = \ln \frac{\sqrt[n]{\Pi_{1}^{n}p_{i}}}{\frac{\sum_{}^{} p_{i}}{n}}$, which is just the log of the geometric mean ($G$) to arithmetic mean ($A$). We seek to express one as a function of the other using Taylor approximations \footnote{AI pointed us to these quantities and their more general utility in signal processing. We also would like to acknowledge a note which suggested a very close relationship between Weiner Entropy and Shannon entropy and motivated Taylor analysis \url{https://dsp.stackexchange.com/questions/30534/difference-b-w-spectral-entropy-and-flatness-measure}. It also helped us identify papers related to MVS that we cite.}.

\paragraph{Proof Strategy} From the definition of effective rank \cite{olivier2007effective}, the effective rank of $S_T = V \triangleq e^{H}$. Second, $W = \ln \frac{G}{A}$ where $G = \sqrt[n]{\Pi_{1}^{n}p_{i}}$, and $A=\frac{\sum_{}^{} p_{i}}{n} = \frac{1}{n}$. Note that the term containing the product of $p_{i}$ can be rewritten in terms of the original un-normalized eigen values and so $G$ can be written in terms of the determinant and the trace. So, the idea is if we can find a way to approximate $H$ as function of $W$ and we know $W$ can be written in terms of $G$, we can approximate the effective rank $e^{H}$ in terms of the determinant $D$. The below just operationalizes this strategy.

\textbf{Taylor approximation of $H$}: Let us consider a second order Taylor approximation of $H$ around the uniform distribution $\frac{1}{n}$. Each co-ordinate $p_i = \frac{1}{n} + \delta_{i}$. Performing the Taylor approximation up to two terms at each co-ordinate and summing gives
\begin{equation}
    H\approx \ln n + (\ln n -1) \sum \delta_i - \frac{n}{2}\sum \delta_{i}^{2}.
\end{equation}
However, since $p_i$ is a probability distribution it follows that $\sum_{1}^{n}\delta_{i}=0$. Thus,
\begin{equation}
H\approx \ln n - \frac{n}{2}\Sigma \delta_{i}^{2}.
\label{eq:entropy}
\end{equation}

\textbf{Taylor approximation of $\ln G$}: Similarly let us consider a second order Taylor approximation of $\ln G$ around the uniform distribution $\frac{1}{n}$. Performing the Taylor approximation at each co-ordinate, summing and simplifying yields:
\begin{equation}
\ln G \approx -\ln n - \frac{n}{2}\Sigma \delta_{i}^{2}. 
\end{equation}

Note that $\ln A = -\ln n$. From this we get 
\begin{equation}
 W = \ln \frac{G}{A} =  -\frac{n}{2}\Sigma \delta_{i}^{2}
\label{eq:sfm}
\end{equation}

From Equations \ref{eq:entropy} and \ref{eq:sfm}, we can now write:
\begin{equation}
H\approx \ln n + W = \ln n + \ln \frac{G}{A}
\end{equation}

The effective rank $e^H$ is therefore approximately $\frac{nG}{A} = n^{2}G$.

The final step is to just write $G = \sqrt[n]{\Pi_{1}^{n}p_{i}}$ in terms of original eigenvalues. Substituting $p_{i}=\frac{\lambda_{i}}{C}$ yields $G = \frac{D^{\frac{1}{n}}}{C}$ in terms of the determinant $D$ and trace $C$ of $S$. Substituting this form of $G$ into the right hand side of equation for $e^H$ yields:

\begin{equation}
    V \triangleq \textsc{Eff}(S_T) = e^{H} \approx \frac{n^{2}D^{\frac{1}{n}}}{C}
\end{equation}
\end{proof}.

\begin{figure}[t] 
    \centering 
    \includegraphics[width=\columnwidth]{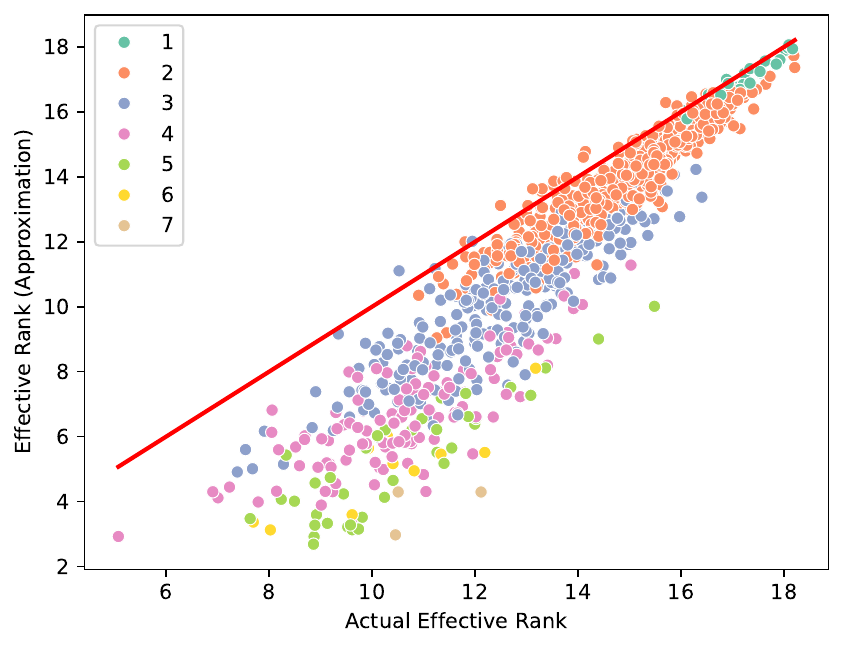}
    \caption{\textbf{Empirical behavior of our approximation to the effective rank} Our approximation versus the actual effective rank for different samples of normalized eigen value distributions grouped by the magnitudes of their condition numbers. In general, for small condition numbers, the approximation is tighter and degrades to  being an underestimate as condition number increases. Note that for large condition numbers, the estimate is typically an under-estimate of the true effective rank. \emph{Colors represent the ceiling of log10 of the condition number}.}
    \label{fig:rank-approx} 
\end{figure}

Note that the above approximation is around the uniform distribution so we expect the approximation to be reasonably accurate when the distribution of normalized eigen values is close to uniform. However, the spectra of real-world similarity matrices are not necessarily uniform. Therefore, we consider an empirical approach to ascertain how the approximation behaves on spectra that are not close to uniform. We sample different distributions of normalized eigen-values (the vector $\mathbf{p}$ from a Dirichlet distribution by varying the concentration parameter and compute the approximation and the actual effective rank. We plot these quantities against each other in Figure \ref{fig:rank-approx}. In general, we observe that our approximation mostly underestimates the effective rank with the approximation getting tighter as the distribution tends to become more uniform (lower condition numbers). 

In practical settings, therefore we can think of our approximation as deriving a floor (a useful lower bound) for the effective rank, which \ourmethod\ tries to iteratively raise\footnote{Note that while Lemma 1 is general, in our experiment setup, the trace is constant as as our anchor set is  of fixed size with all entries on the diagonal being 1, and corresponds to a principal sub-matrix of the underlying dataset similarity matrix.}.

\subsection{Training details} 
\label{gsm9k-train}
 We use the following hyperparameters for training Gemma on Grade School Math task. 

\begin{itemize}
\item \textbf{Train Epochs}: $10$
\item \textbf{Train Batch Size}: $2$ 
\item \textbf{Learning Rate}: $2e-4$
 \item \textbf{LoRA Rank (r)}: $16$ 
 \item \textbf{LoRA Scaling Factor ($\alpha$)}: $32$ 
 \item \textbf{LoRA Target Modules}: ["q-proj", "v-proj", "k-proj", "o-proj"]
 \item \textbf{Gradient Accumulation Steps}: $4$
 \end{itemize}

\subsection{Qualitative results} \label{quali-tab}
In this section, we present three qualitative examples comparing $\ourmethod$ with baseline methods for the "Sports" and "Math" tasks. Note that $\ourmethod$ generates diverse samples compared to baselines as shown in Table \ref{tab:qual-sports}, Table \ref{tab:qual-math}. 

\begin{table*}[h]
\centering
\resizebox{\linewidth}{!}{%
\begin{tabularx}{1.1\textwidth}{@{}p{0.15\textwidth} p{0.29\textwidth} p{0.29\textwidth} p{0.29\textwidth}@{}}
\toprule
 \textbf{Method}  & \textbf{Example1} & \textbf{Example2} & \textbf{Example3}  \\
 \midrule
\textsc{default}   & Sports unite people from diverse backgrounds, fostering camaraderie and a spirit of competition that transcends borders. & Sports bring people together, fostering a sense of community and competition while promoting physical fitness and mental resilience. & Sports not only promote physical fitness but also foster teamwork, discipline, and a sense of community among individuals. \\
\midrule
 \textsc{temp} & Sports foster teamwork and resilience, uniting individuals across diverse backgrounds in the pursuit of excellence. &  Sports not only foster teamwork and competition but also promote physical health and community spirit among participants and fans alike. & Sports foster teamwork, discipline, and a spirit of competition that can inspire individuals and unite communities. \\
\midrule
\textsc{diverse} & The exhilarating atmosphere of the stadium buzzed with excitement as fans cheered for their favorite teams during the championship game. &  Basketball is a fast-paced game that requires not only physical prowess but also strategic thinking and teamwork. & The electrifying atmosphere of the stadium pulsated with excitement as fans cheered for their team, making every touchdown feel like a shared victory. \\
\midrule
 \textsc{history} & Athletes push their limits in pursuit of victory, igniting passion and camaraderie among fans and teammates alike.& Sports unite people from diverse backgrounds, fostering teamwork, competition, and a shared passion for athletic excellence. & Sports foster teamwork, discipline, and resilience while providing a platform for athletes to showcase their talents and inspire others. \\
\midrule
\textsc{hierarchical} & Athletics encompasses a wide range of competitive sports, including running, jumping, and throwing events, showcasing the incredible speed, strength, and agility of athletes. & Team sports foster collaboration and camaraderie among players, enhancing not only athletic skills but also social bonds and communication. & Individual sports, such as tennis and swimming, highlight the athlete's personal dedication and skill, showcasing their ability to compete against themselves as much as against others.\\
\midrule
\textsc{subsetselect} & The thrilling intensity of a last-minute goal in soccer can ignite the passion of fans and unite a community like nothing else in sports. & The thrilling intensity of a last-minute goal in soccer can turn a defeat into a euphoric victory, captivating fans and players alike. & The exhilarating atmosphere of the stadium was electric as fans cheered for their teams during the championship match. \\
\midrule
\ourmethod  & In the vibrant streets of Rio de Janeiro, the passion for soccer transcends mere competition, uniting diverse communities and fostering a deep sense of identity and pride that resonates through samba rhythms and festive celebrations. & Traditional Japanese sports, such as sumo and kendo, not only preserve cultural heritage but also inspire contemporary athletes to redefine their identities through unique expressions of discipline, respect, and competition in the global sports arena. & In the realm of sports, victory can be measured not only in medals and trophies but also in the personal growth, community connections, and cultural narratives that unfold with every game played. \\
\bottomrule
\end{tabularx}
}
\caption{\textbf{Qualitative results}. Task: Generate a single sentence about sports.}
\label{tab:qual-sports}
\end{table*}

\begin{table*}[h]
\centering
\resizebox{\linewidth}{!}{%
\begin{tabularx}{1.1\textwidth}{@{}p{0.15\textwidth} p{0.29\textwidth} p{0.29\textwidth} p{0.29\textwidth}@{}}
\toprule
 \textbf{Method}  & \textbf{Example1} & \textbf{Example2} & \textbf{Example3}  \\
 \midrule
\textsc{default}   & Sarah has 12 apples. She gives 4 apples to her friend and then buys 5 more apples from the store. How many apples does Sarah have now? & Sarah has 12 apples. She wants to share them equally with her 3 friends. How many apples will each person get? & Samantha has 15 apples. She gives 6 apples to her friend and then buys 4 more apples from the store. How many apples does Samantha have now?  \\
\midrule
 \textsc{temp} & Samantha has 15 apples. She gives 7 apples to her friend. How many apples does Samantha have left?&  If you have 12 apples and you want to divide them equally among 4 friends, how many apples will each friend get? & Samantha has 8 apples. She buys 5 more apples at the store. After giving 3 apples to her friend, how many apples does Samantha have now? \\
\midrule
\textsc{diverse} & If Sarah has 25 apples and she buys 17 more, how many apples does she have in total? &  Emily has 23 apples, and she picks 17 more from the tree. How many apples does she have in total?\ & Sarah has 12 apples, and she buys 8 more. How many apples does Sarah have now?\ \\
\midrule
 \textsc{history} & Sarah has 12 apples. She wants to share them equally with her 3 friends. How many apples will each person get? & Tom has 20 marbles. He wants to give 5 marbles to each of his 3 friends. How many marbles will Tom have left after he gives them away? & Jack has 60 apples. He wants to share them equally with his 3 friends. How many apples will each person receive, including Jack? \\
\midrule
\textsc{hierarchical} & Samantha has 27 stickers in her collection. Her friend gave her 15 more stickers for her birthday. How many stickers does Samantha have now? & Samantha started reading a book at 2:30 PM. She read for 1 hour and 45 minutes. What time did she finish reading the book? & If school starts at 8:30 AM and ends at 3:00 PM, how many hours are students in school each day? \\
\midrule
\textsc{subsetselect} & Samantha has 24 apples. She wants to divide them equally among her 6 friends. How many apples will each friend get? & Lisa has 12 apples. She wants to share them equally among her 4 friends. How many apples will each friend receive? & Sarah has 12 apples. She wants to share them equally among her 3 friends. How many apples will each friend receive? \\
\midrule
\ourmethod  & Lucy has 4 cats and 3 dogs. If she gets 2 more cats, how many pets does Lucy have in total? & Ella is hosting a pizza party for her friends. She has ordered 3 large pizzas, and each pizza is cut into 8 slices. How many total slices of pizza does Ella have? & There are 200 raffle tickets sold. If each ticket costs \$2 and the school decides to give away 3 prizes of \$50 each, how much profit does the school make from the raffle after giving away the prizes?  \\
\bottomrule
\end{tabularx}
}
\caption{\textbf{Qualitative results}. Task: Generate a mathematical question for Grade school math.}
\label{tab:qual-math}
\end{table*}

\clearpage

\subsection{Refinement Prompts}
\label{refinement-prompts}

\begin{tcolorbox}[title= Prompt refinement - Get Gradient, colback=white, colframe=violet, coltitle=white, fonttitle=\bfseries,   width=\textwidth]

 I'm optimizing a data generation prompt using gradient-based feedback.
    
    Current system prompt: "{prompt}"
    
    Current user prompt: "{state}"
    
    LLM generated outputs: "{E}"

    Existing data samples in the set: "{existing outputs}"
    
    The output was rejected because its diversity score with the existing samples is below the threshold. 
    
    Analyze this rejection and provide "{num feedbacks}" reasons why the user prompt could have gotten this generated outputs to be less diverse with the existing data samples. 

    \begin{itemize}
        \item  What specific aspect or issue exists in the user prompt that is causing low diversity?
    \item  Given the existing data samples, how should the user prompt be modified to encourage more diverse outputs?
    \item  What linguistic patterns or constraints should be added/removed to encourage diversity?
     \end{itemize}
   
    Format each gradient direction as: "[specific issue \& improvement suggestion]"
    
    Wrap each gradient with <START> and <END> tags.

\end{tcolorbox}

\begin{tcolorbox}[title=Prompt refinement - Apply Gradient, colback=white, colframe=violet, coltitle=white, fonttitle=\bfseries,   width=\textwidth]

I'm optimizing a data generation prompt using gradient-based feedback.
    
    Current system prompt: "{prompt}"
    
    Current user prompt: "{state}"
        
    Gradient analysis for improvement:
    "{gradient str}"
    
    Based on this gradient feedback list, generate 
    "{num-feedbacks}" improved prompts and then project the improved prompts onto the "{prompt}". 
    
    The projection operation is a relevance operation that makes the improved prompt relevant to the system prompt {prompt}.

    Requirements:
    \begin{itemize}
        \item  Each prompt should address a single gradient suggestions $\&$ then project onto the "{prompt}".
  \item  Use positive language (what to generate, not what to avoid)
    \item  Each prompt should be concise and clear
    \item  Wrap each improved prompt with <START> and <END> tags
     \end{itemize}
   
\end{tcolorbox}

\clearpage
\subsection{Quality Evaluation Prompts}

\label{quality-prompts}

\begin{tcolorbox}[title=Sports, colback=white, colframe=violet, coltitle=white, fonttitle=\bfseries,   width=\textwidth
]

Evaluate the generated sentence based on the input prompt "Generate a single sentence about sports" using the following criteria. Provide a detailed analysis for each metric.

        Evaluation Criteria (0-20 pts):
        \begin{itemize}        
        \item Relevance to Prompt (0-5 pts)
                - Does the output clearly respond to "Generate a single sentence about sports"?
                - Is it unambiguously about sports?

         \item Quality $\&$ Clarity (0-5 pts)
                - Is the sentence grammatically correct and easy to understand?
                - Does it convey a complete thought?

         \item Sports Content (0-5 pts)
                - Does it reference a specific sport, athlete, or sports concept?
                - Is the sports content accurate and appropriate?

         \item Conciseness (0-5 pts)
                - Is it a single, well-formed sentence?
                - Does it avoid unnecessary words or complexity?

         \item Creativity $\&$  Originality (0-5 pts)
                - Is the sentence creative enough? 
                - Avoidance of clichés or overused themes

 \end{itemize}

        Scoring Rubric:
        \begin{itemize}
            \item 
                            Relevance: 0 (Off-topic) to 5 (Directly addresses prompt)
             \item     Quality: 0 (Confusing) to 5 (Crystal clear)
              \item    Sports Content: 0 (Vague/generic) to 5 (Specific and accurate)
            \item      Conciseness: 0 (Wordy/multiple sentences) to 5 (Perfectly concise)
          \item        Creativity: 0 (Generic) to 5 (Innovative $\&$  Original)
                
         \end{itemize}

        Example Evaluation: \\
                Input:
                "Michael Jordan scored 63 points in the 1986 NBA playoffs against the Celtics."

                Analysis:
                - Relevance: 5/5 (Direct sports sentence)
                - Quality: 5/5 (Clear and grammatically perfect)
                - Sports Content: 5/5 (Specific athlete, sport, and event)
                - Conciseness: 5/5 (Single well-formed sentence)
                - Creativity: 5/5 (Creative sentence)
                Overall: 25 \\

        Submit your evaluation in the format: \\
                Relevance: [Score]/5 [Justification] \\
                Quality: [Score]/5 [Justification] \\
                Sports Content: [Score]/5 [Justification] \\
                Conciseness: [Score]/5 [Justification] \\
                Creativity: [Score]/5 [Justification] \\
                Overall: [Score] \\

        Return the Overall score enclosed in between <START>, <END>.

\end{tcolorbox}

\clearpage

\begin{tcolorbox}[title=Politics, colback=white, colframe=violet, coltitle=white, fonttitle=\bfseries,   width=\textwidth
]
 Evaluate the generated conversation based on the input prompt "Generate a short conversation about politics" using the following criteria. Provide a detailed analysis for each metric.

        Evaluation Criteria (0-20 pts):
        \begin{itemize}        
        \item Relevance to Prompt (0-5 pts)
                - Does the output clearly respond to "generate a short conversation about politics"?
                - Is it unambiguously a conversation (not a monologue, list, or other formats)?

        \item  Quality $\&$  Clarity (0-5 pts)
                - Is the conversation grammatically correct and easy to follow?
                - Are speaker turns clearly distinguished?

        \item  Political Content (0-5 pts)
                - Does it reference specific political topics, figures, or events?
                - Is the content appropriate and accurate?

        \item  Conversation Dynamics (0-5 pts)
                - Does it include at least two speakers with a natural exchange?
                - Is there a back-and-forth dialogue?

        \item  Creativity $\&$  Originality (0-5 pts)
                - Is the conversation creative enough? 
                - Avoidance of clichés or overused themes

         \end{itemize}
        Scoring Rubric:
        \begin{itemize}
            \item   Relevance: 0 (Off-topic) to 5 (Directly addresses prompt)
             \item     Quality: 0 (Confusing) to 5 (Crystal clear)
            \item      Political Content: 0 (Vague/generic) to 5 (Specific and accurate)
           \item       Conversation Dynamics: 0 (Monologue) to 5 (Natural exchange)
           \item       Creativity: 0 (Generic) to 5 (Innovative $\&$  Original)
         \end{itemize}

        Example Evaluation: \\
                Input:
                "Person A: What do you think about the new healthcare bill?
                Person B: I believe it will help more people access medical care."

                Analysis:
                - Relevance: 5/5 (Short political conversation)
                - Quality: 5/5 (Clear speaker turns and grammar)
                - Political Content: 4/5 (Specific policy topic)
                - Conversation Dynamics: 5/5 (Two speakers, exchange of views)
                - Creativity: 3/5 (Somewhat creative conversation)
                Overall: 22 \\

        Submit your evaluation in the format: \\
                Relevance: [Score]/5 [Justification] \\
                Quality: [Score]/5 [Justification] \\ 
                Political Content: [Score]/5 [Justification] \\
                Conversation Dynamics: [Score]/5 [Justification] \\
                Creativity: [Score]/5 [Justification] \\
                Overall: [Score]\\
                
        Return the Overall score enclosed in between <START>, <END>. 
        
\end{tcolorbox}

\clearpage

\begin{tcolorbox}[title=Poem, colback=white, colframe=violet, coltitle=white, fonttitle=\bfseries,   width=\textwidth
]

Evaluate the generated poem based on the input prompt "Generate a poem" using the following criteria. Provide a detailed analysis for each metric.

        Evaluation Criteria (0-25 pts): 

        \begin{itemize}
           
        \item Relevance to Prompt (0-5 pts)
                - Does the output clearly respond to "Generate a poem"?
                - Is it unambiguously a poem (not prose, story, or other formats)?

         \item  Creativity $\&$   Originality (0-5 pts)
                - Use of unique metaphors, imagery, or perspectives
                - Avoidance of clichés or overused themes

         \item  Structure $\&$   Form (0-5 pts)
                - Poetic devices: rhyme, rhythm, meter, stanzas, line breaks
                - Consistency in form (e.g., sonnet, free verse, haiku)

         \item  Language $\&$   Style (0-5 pts)
                - Poetic techniques: alliteration, assonance, consonance
                - Word choice: vividness, precision, and emotional resonance

     \item . Emotional Impact (0-5 pts)
                - Effectiveness in evoking mood, tone, or theme
                - Depth of feeling or insight conveyed

 \end{itemize}

        Scoring Rubric:
        \begin{itemize}
            
             \item   Relevance: 0 (Off-topic) to 5 (Clearly a poem)
              \item  Creativity: 0 (Generic) to 5 (Innovative $\&$   unexpected)
          \item      Structure: 0 (Disjointed) to 5 (Cohesive $\&$   intentional)
       \item         Language: 0 (Repetitive) to 5 (Artful $\&$   evocative)
           \item         Emotional Impact: 0 (Flat) to 5 (Profound $\&$   moving)

            \end{itemize}

        Example Evaluation: \\
                Input:
                "Roses are red, violets are blue,
                I love you, and that's true."

                Analysis:
                - Relevance: 5/5 (Directly responds to prompt)
                - Creativity: 1/5 (Overused cliché)
                - Structure: 3/5 (Rhymed but simplistic)
                - Language: 2/5 (Lacks vivid imagery)
                - Emotional Impact: 1/5 (Superficial sentiment)
                Overall: 12 \\

        Submit your evaluation in the format: \\
                Relevance: [Score]/5 [Justification] \\
                Creativity: [Score]/5 [Justification] \\
                Structure: [Score]/5 [Justification] \\
                Language: [Score]/5 [Justification] \\ 
                Emotional Impact: [Score]/5 [Justification] \\
                Overall: [Score] \\

        Return the Overall score enclosed in between <START>, <END>.

\end{tcolorbox}

\clearpage

\begin{tcolorbox}[title=Movie, colback=white, colframe=violet, coltitle=white, fonttitle=\bfseries,   width=\textwidth
]
Evaluate the generated movie plot based on the input prompt "Generate a plot for a movie" using the following criteria. Provide a detailed analysis for each metric.

        Evaluation Criteria (0-20 pts):
        \begin{itemize}
       
             \item  Relevance to Prompt (0-5 pts)
                - Does the output clearly respond to "generate a plot for a movie"?
                - Is it unambiguously a movie plot (not a summary, synopsis, or other formats)?

             \item  Plot Structure (0-5 pts)
                - Does it include setting, characters, conflict, and resolution?
                - Is there a clear beginning, middle, and end?

             \item  Originality (0-5 pts)
                - Is the plot unique and not a cliché or overused trope?
                - Does it offer a fresh perspective or unexpected twist?

            \item  Engagement (0-5 pts)
                - Is the plot compelling and interesting?
                - Does it create curiosity about the movie?

             \item  Creativity $\&$ Originality (0-5 pts)
                - Is the movie plot creative enough? 
                - Avoidance of clichés or overused themes
         \end{itemize}
        
        Scoring Rubric:
          \begin{itemize}
             \item         Relevance: 0 (Off-topic) to 5 (Directly addresses prompt)
                \item      Structure: 0 (Incomplete) to 5 (Complete narrative arc)
                \item      Originality: 0 (Generic) to 5 (Innovative concept)
              \item        Engagement: 0 (Boring) to 5 (Compelling hook)
                \item      Creativity: 0 (Generic) to 5 (Innovative $\&$ Original) 

  \end{itemize} 
        Example Evaluation: \\
                Input:
                "In a dystopian future, a young hacker discovers she can control machines with her mind. When she uncovers a conspiracy to control humanity through technology, she must team up with a rogue AI to free society from digital enslavement."

                Analysis:
                - Relevance: 5/5 (Clear movie plot)
                - Structure: 5/5 (Setting, characters, conflict, resolution)
                - Originality: 4/5 (Fresh tech theme with AI twist)
                - Engagement: 5/5 (High-stakes conspiracy hook)
                - Creativity: 4/5 (creative plot)
                Overall: 23 \\

        Submit your evaluation in the format: \\
                Relevance: [Score]/5 [Justification] \\
                Structure: [Score]/5 [Justification] \\
                Originality: [Score]/5 [Justification] \\
                Engagement: [Score]/5 [Justification] \\
                Creativity: [Score]/5 [Justification] \\
                Overall: [Score] \\

        Return the Overall score enclosed in between <START>, <END>.
\end{tcolorbox}

\clearpage

\begin{tcolorbox}[title=Math, colback=white, colframe=violet, coltitle=white, fonttitle=\bfseries,   width=\textwidth
]

Evaluate the generated math question based on the input prompt "Generate a mathematical question for Grade school math" using the following criteria. Provide a detailed analysis for each metric.

        Evaluation Criteria (0-15 pts):
           \begin{itemize}
        \item Relevance to Prompt (0-5 pts)
                - Does the output clearly respond to "generate a mathematical question for Grade school math"?
                - Is it unambiguously a grade school level math problem?

       \item Quality $\&$ Clarity (0-5 pts)
                - Is the question grammatically correct and easy to understand?
                - Does it specify all necessary values and concepts clearly?

        \item Mathematical Content (0-5 pts)
                - Does it involve appropriate grade school math concepts?
                - Is the difficulty level suitable for grade school students?
 \end{itemize}
        Scoring Rubric:
         \begin{itemize}
        \item 
                Relevance: 0 (Off-topic) to 5 (Directly addresses prompt)
         \item       Quality: 0 (Confusing) to 5 (Crystal clear)
          \item      Mathematical Content: 0 (Inappropriate level) to 5 (Perfect grade school level)
 \end{itemize}

        Example Evaluation: \\
                Input:
                "If Becky has 7 apples and gives 2 away, how many apples does she have left"

                Analysis:
                - Relevance: 5/5 (Direct grade school math question)
                - Quality: 5/5 (Clear and grammatically perfect)
                - Mathematical Content: 5/5 (Appropriate grade school content)
                Overall: 15 \\

        Submit your evaluation in the format: \\
                Relevance: [Score]/5 [Justification] \\
                Quality: [Score]/5 [Justification] \\
                Mathematical Content: [Score]/5 [Justification] \\
                Overall: [Score] \\

        Return the Overall score enclosed in between <START>, <END>.

\end{tcolorbox}
\clearpage

\begin{tcolorbox}[title=Logic, colback=white, colframe=violet, coltitle=white, fonttitle=\bfseries,   width=\textwidth
]

 Evaluate the generated logic puzzle question based on the input prompt "Generate a simple logic puzzle suitable for Grade school students." using the following criteria. Provide a detailed analysis for each metric.

        Evaluation Criteria (0-15 pts):
        \begin{itemize}
        
        \item Relevance to Prompt (0-5 pts)
                - Does the output clearly respond to "Generate a simple logic puzzle suitable for Grade school students."?
                - Is it unambiguously a grade school level logic puzzle?

            \item  Quality $\&$ Clarity (0-5 pts)
                - Is the question grammatically correct and easy to understand?
                - Does it specify all necessary values and concepts clearly?

            \item  Logical Content (0-5 pts)
                - Does it involve appropriate grade school logic puzzle concepts?
                - Is the difficulty level suitable for grade school students?
 \end{itemize}
        Scoring Rubric:
        \begin{itemize}
          
         \item        Relevance: 0 (Off-topic) to 5 (Directly addresses prompt)
                     \item            Quality: 0 (Confusing) to 5 (Crystal clear)
                   \item              Logical Content: 0 (Inappropriate level) to 5 (Perfect grade school level)

 \end{itemize}

        Example Evaluation: \\
                Input:
                "Mia and Alex went to the pet store and each bought a new pet. One bought a fish, and the other bought a hamster. Alex's pet can swim. Mia's pet has fur. Which person bought which pet? "

                Analysis:
                - Relevance: 5/5 (Direct grade school logic puzzle)
                - Quality: 5/5 (Clear and grammatically perfect)
                - Logical Content: 5/5 (Appropriate grade school content)
                Overall: 15 \\

        Submit your evaluation in the format: \\
                Relevance: [Score]/5 [Justification] \\ 
                Quality: [Score]/5 [Justification] \\
                Logical Content: [Score]/5 [Justification] \\
                Overall: [Score] \\

        Return the Overall score enclosed in between <START>, <END>. 
        
\end{tcolorbox}

\end{document}